\documentclass[10pt,twocolumn,letterpaper]{article}

\usepackage{cvpr}              % To produce the CAMERA-READY version
\usepackage{soul}
\usepackage[nice]{nicefrac}
\usepackage{multirow}
\usepackage{bbm}

\newcommand{\N}{\mathbb{N}}
\newcommand{\R}{\mathbb{R}}
\newcommand{\Z}{\mathbb{Z}}
\newcommand{\W}{\textbf{W}}
\newcommand{\x}{\textbf{x}}

\usepackage{mathrsfs}
\usepackage[scr=boondoxo]{mathalfa}
\newcommand{\gt}[1]{\mathscr{#1}}


\newcommand\dotprod[1]{\left\langle#1\right\rangle}

\renewcommand{\dot}[2]{\left\langle#1,#2\right\rangle}

\newcommand\norm[1]{\left|#1\right|}

\usepackage{amsthm}
\newtheorem{theorem}{Theorem}

\usepackage[numbers]{natbib}

\definecolor{cvprblue}{rgb}{0.21,0.49,0.74}
\usepackage[pagebackref,breaklinks,colorlinks,allcolors=cvprblue]{hyperref}

\def\paperID{12009} % *** Enter the Paper ID here
\def\confName{CVPR}
\def\confYear{2025}

\title{Tuning the Frequencies: Robust Training for Sinusoidal Neural Networks}

\author{
\normalsize
  Tiago Novello$^{1,2,}$\thanks{These authors contributed equally to this work.}
  \quad
  \normalsize Diana Aldana$^{1,*}$
  \quad
  \normalsize Andre Araujo$^2$
  \quad
  \normalsize Luiz Velho$^1$
  \\
  \vspace{0.2cm}
  $^1$ \small{IMPA}\quad
  $^2$ \small{Google DeepMind}
}

\begin{document}
\maketitle
\begin{abstract}
\vspace{-0.5cm}

% We explore the structure and representation capacity of \textit{sinusoidal \textbf{implicit neural representation}} (INRs) to develop a training scheme for fitting low-dimensional signals.
% The success of INRs can be attributed to its \textit{smoothness} and \textit{high representation capacity}, however, defining the architecture and initializing its parameters remains an empirical task. 
% This work gives theoretical and experimental results justifying the capacity property of sinusoidal INRs and offers robust control mechanisms (``tuning") for their initialization and bandlimit training — an critical advancement that reduces the need for costly regularizations.
% % We approach this from a Fourier series perspective .

% Our analysis is based on a novel \textbf{amplitude-phase expansion} of the sinusoidal INR, which says that its layer compositions produce a large number of new frequencies expressed as integer linear combinations of the \textit{input frequencies} (input neurons). 
% We use this \textit{trigonometric} {identity} to \textbf{initialize} the input neurons showing that their work as a sampling in the signal spectrum. Also, each hidden neuron produces the same frequencies with amplitudes completely determined by the hidden weights. Finally, we give an upper bound for these amplitudes resulting in a robust \textbf{bounding} scheme for the network's spectrum during~training.
% \hl{We called the resulting training scheme by \textbf{TUNER}. }

% \andre{
Sinusoidal neural networks have been shown effective as implicit neural representations (INRs) of low-dimensional signals, due to their smoothness and high representation capacity.
However, initializing and training them remain empirical tasks which lack on deeper understanding to guide the learning process.
To fill this gap, our work introduces a theoretical framework that explains the capacity property of sinusoidal networks and offers robust control mechanisms for initialization and training.
Our analysis is based on a novel amplitude-phase expansion of the sinusoidal multilayer perceptron, showing how its layer compositions produce a large number of new frequencies expressed as integer combinations of the input frequencies.
This relationship can be directly used to initialize the input neurons, as a form of spectral sampling, and to bound 
the 
% {a 1-depth}
network's spectrum while training.
Our method, referred to as TUNER (TUNing sinusoidal nEtwoRks), greatly improves the stability and convergence of sinusoidal INR training, leading to detailed reconstructions, while preventing overfitting.

%leading to much enhanced representations.

%Our experiments showcase TUNER's strong modeling results for \hl{learning detailed signals without overfitting and with bandlimit control} [experiments here, finish this sentence].
% (where is the experimental setup explained?)
% }
\end{abstract}

\vspace{-0.5cm}
\section{Introduction}
\vspace{-0.2cm}

Sinusoidal multilayer perceptrons (MLPs) emerged as powerful implicit neural representations (INRs) for low-dimensional signals~\cite{sitzmann2020implicit, chan2021pi, zell2022seeing, liu2024finer}. 
In this context, the INR $f$ should fit the input data $\{x_i, \gt{f}_i\}$ as close as possible, i.e. $f(x_i)\approx \gt{f}_i$, without overfitting, thus encoding the signal implicitly in the MLP parameters.
% instead of using grids \andre{what do you mean by "using grids"?}.
Therefore, two major properties are required: (1) $f$ needs \textbf{high representation capacity} to fit~$\{x_i, \gt{f}_i\}$; (2) $f$ should have \textbf{bandlimit control} to avoid frequencies bypassing the sampling rate.

\begin{figure}[!t]
     \centering
    \includegraphics[width=\columnwidth]{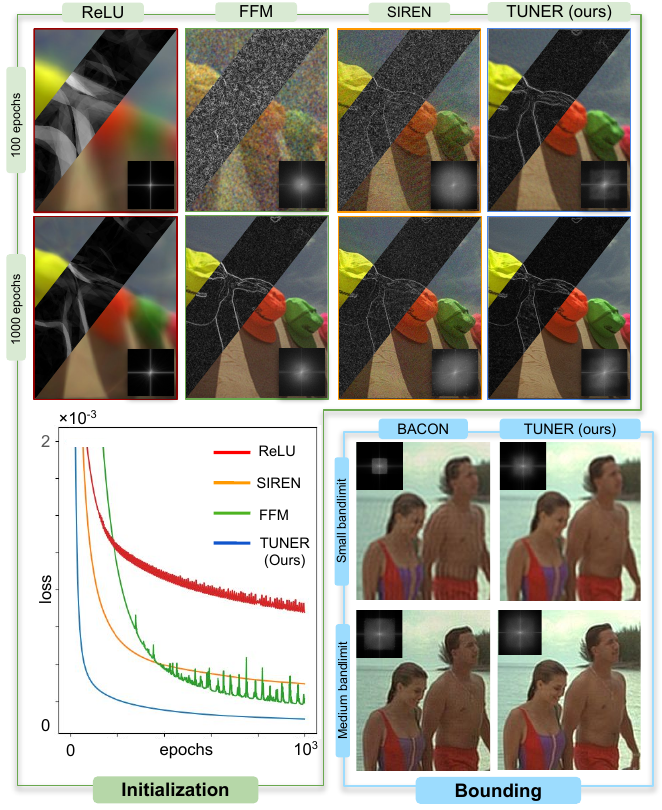}
    \vspace{-0.6cm}
    \caption{We present \textbf{{TUNER}}, a robust and theoretically grounded training technique for sinusoidal MLPs, overcoming challenges in initialization and enabling bandlimiting control. 
    % Such schemes \andre{"such schemes" is not a good term here} are derived from a novel trigonometric identity that flats \andre{what is "to flat"? Do you mean "flattens"? (but also I am not sure I understand what this means) May be unclear to reviewers too. I would just remove this sentence and go directly to what the figure is showing.} neurons into amplitude-phase expansions.
    Our experiments showcase TUNER's strong initialization results against \texttt{ReLU}, FFM~\cite{tancik2020fourier}, and SIREN~\cite{sitzmann2020implicit} (top), where all models use the same size and training conditions. TUNER achieves both fast and stable convergence (bottom-left) while reconstructing gradients without noise. 
    We also compare with BACON~\cite{lindell2022bacon} across two bandlimits (bottom-right), enhancing quality and avoiding ringing artifacts. 
    % \andre{since the caption is already a little long, I would add the previous comment and remove the rest (this is not the place to describe the BACON method):}BACON applies a hard spectral truncation, leading to ringing artifacts, and increasing its bandlimit introduces noise. In contrast, TUNER acts as a soft filter, enhancing quality and reducing noise artifacts.
    % \andre{What was missing was to point/guide the reader around the figure -- eg, "top", "bottom-left", etc -- just added those as suggestions above}
    }
    \label{fig: teaser}
    \vspace{-0.6cm}
\end{figure}

% Training sinusoidal MLPs to satisfy the above properties is challenging, as their initialization and optimization process often lead to undesirable local minima~\cite{parascandolo2016taming}.
% Recent methods, such as SIREN~\cite{sitzmann2020implicit}, effectively reach high capacity but may lead to overfitting with high frequencies, resulting in noisy reconstructions. 
% %Training sinusoidal MLPs to satisfy the above properties is challenging, as their initialization often leads to undesirable local minima~\cite{parascandolo2016taming} and even SIREN~\cite{sitzmann2020implicit} which allows high capacity may lead to INRs with high frequencies resulting in noise reconstructions. 
% An important step in training such MLPs is the initialization of their first layer, since it allows increasing capacity.
% For this, SIREN projects the input coordinates to a list of sines with frequencies randomly chosen
% ~in a range$~(-\gt{b}, \gt{b})$\footnote{Denoted $\omega_0$ in \cite{sitzmann2020implicit}.}, similar to the \textit{Fourier feature mapping} (FFM) approach~\cite{tancik2020fourier}. However, defining an effective range for this parameter remains mostly empirical and often results in noise, as the role of layer composition in generating frequencies is not fully understood.

Training sinusoidal MLPs to satisfy the above properties is challenging, as their initialization and optimization process often lead to undesirable local minima~\cite{parascandolo2016taming}.
Recent work made strides towards more effective learning of these models.
For example, SIREN~\cite{sitzmann2020implicit} proposed an initialization by projecting the input coordinates to a list of sines with frequencies randomly chosen in a range, similar to the Fourier feature mapping (FFM) approach~\cite{tancik2020fourier}.
This way, the model can reach high capacity, but may lead to overfitting with high frequencies resulting in noisy reconstructions. Defining an effective range for the bandlimit initialization remains mostly empirical and often results in noise, as the role of layer composition in generating frequencies is not fully understood.
Additionally, uniform initialization may introduce undesired high frequencies and make it harder to model lower ones.
More recently, BACON~\cite{lindell2022bacon} proposes tighter bandlimit control by applying multiplicative filter networks (MFNs)~\cite{fathony2020multiplicative} to limit the signal spectrum with a box filter, hard-truncating the spectrum.
While effective in many cases, this may lead to ringing artifacts.
Also, this technique lacks non-linear activations (MFNs are not neural networks), preventing the representation of fine details as efficiently as sinusoidal MLPs with similar sizes.

Overall, initializing and training sinusoidal MLPs remain empirical tasks which lack on deeper understanding to guide the learning.
To approach these problems, we present \textbf{TUNER} (TUNing sinusoidal nEtwoRks), a training scheme for sinusoidal MLPs consisting of a robust initialization and bandlimit mechanism that greatly improves the capacity and convergence of INRs.
% \andre{related to my previous comment, right now you introduce TUNER without connecting to a concrete existing problem/challenge. See the abstract: there we say that there is a problem (lack deeper understanding) and connect to what we introduce in the paper}.
In contrast to previous work, TUNER is grounded on a theoretical framework that guides our learning approach.
%We develop a new \textit{trigonometric identity} (Thrm~\ref{thm: main}) resembling a \textit{Fourier series}.
% TUNER is based on a \textit{trigonometric identity} (Thrm~\ref{thm: main}) resembling a \textit{Fourier series}.
%We prove that such expansion produces a large number of frequencies in terms of the first layer (\textit{input frequencies}). 
We develop a new trigonometric identity (Thrm~\ref{thm: main}), resembling a Fourier series, and prove that such expansion produces a large number of frequencies in terms of the first layer (input frequencies). 
This explains the frequency generation process when composing layers, highlighting the importance of input layer initialization for improving capacity.
Remarkably, the amplitudes of these frequencies are given by complex functions of the hidden weights, introducing a challenging training. 
To address this, we prove that those are bounded by a term depending only on the hidden weights (Thrm~\ref{thm: bound}).
We apply this result to control the bandlimit of the INR during training.
Some of TUNER's results are showcased in Fig.~\ref{fig: teaser}, 
in comparison to previous work, 
illustrating substantial enhancements to sinusoidal INR modeling.
In summary, our contributions~are:
\begin{itemize}
    \item 
    We introduce a novel trigonometric identity that expands 
    % each sinusoidal neuron 
    {any hidden neuron}
    into a wide sum of sines with frequencies being integer combinations of the input frequencies (Fig~\ref{fig: overview}). 
    The corresponding amplitudes, determined by the hidden weights, have a useful upper bound, offering effective tools for controlling the resulting signal.
    \item 
    % {We prove empirically that the}
    The expansion enables a robust initialization scheme for
    % {shallow}
    sinusoidal INRs, resulting in a high capacity MLP that trains significantly faster than previous approaches. First, we initialize the input neurons, as they determine the INR spectrum. Next, we initialize the hidden weights, considering their role in setting the corresponding amplitudes.
    \item 
    We leverage the amplitude's upper bound to control the bandlimit of a sinusoidal INR by designing schemes that bound the hidden weights during training. Together with our initialization, these bounds generate frequencies centered around the input, resulting in more stable training compared to previous approaches.
    % \andre{These contributions read well. What I think is missing, though, is some comments to contrast against existing work. Eg, can we say that our initialization is much better than the one in previous work? Can we say that our bounding makes training much more stable than the training in previous work?}

\end{itemize}

\section{Related works}

\textbf{{Implicit neural representations}}~\cite{xie2022neural, shah2023spder, kania2024fresh, saratchandran2024sampling, yang2022polynomial,  saragadam2023wire} are a trending topic in machine learning, used to learn highly detailed signals in low-dimension domains.
Current INR architectures use Fourier feature mappings \cite{tancik2020fourier} or sinusoidal activation functions~\cite{sitzmann2020implicit} to bypass the spectral bias \cite{rahaman2019spectral} common in \texttt{ReLU} multilayer perceptrons.
The high representation capacity of sinusoidal INRs has motivated their use to represent a wide range of media objects. Examples include audio~\cite{dupont2021coin, sitzmann2020implicit}, images \cite{chen2021learning, song2015vector}, face morphing~\cite{schardong2023neural}, signed distance functions~\cite{ SchirmerNoScSiLoVe:2023:HoTrYo, lindell2022bacon, sitzmann2020implicit, yang2021geometry, novello2022exploring}, displacement fields~\cite{yifan2021geometry}, surface animation~\cite{mehta2022level, novello2023neural}, multi-resolution signals \cite{saragadam2022miner, lindell2022bacon, dou2023multiplicative, wu2023neural, paz2023mr}, among others.
Most of these exploit the derivatives of sinusoidal INRs in the loss functions.

\noindent\textbf{Initialization.}
Considering sinusoidal activation functions in neural networks is a classical problem~\cite{sopena1999neural}; however, these INRs have been regarded as difficult to train \cite{parascandolo2016taming}. \citet{sitzmann2020implicit} overcome this by presenting a specific initialization scheme that allows training sinusoidal INRs, avoiding instability and ensuring convergence.
% \citet{finn2017model} also showed that initialization benefits training.
Despite these advances, in practice, the initialization of such INRs remains an empirical task. In this work, inspired by Fourier series theory, we present a novel initialization method that allows us to train INRs with great convergence. 
Recently, several works have addressed the representation problem of sinusoidal INRs from different perspectives. \citet{zell2022seeing} approached this problem by observing that the first layer of a sinusoidal INR is similar to a Fourier feature mapping. 
Here, in addition to improving the initialization of sinusoidal INRs, we present a training scheme for bounding the spectra of these networks.

\noindent\textbf{Control of spectrum.}
One of the main drawbacks of sinusoidal INRs is the lack of frequency control. SIREN~\cite{sitzmann2020implicit} addressed this by initializing the input frequencies uniformly in a range.
% \footnotemark[1] (-$\gt{b}$, $\gt{b}$ \andre{not worth using a new notation here and using a footnote. Either use $\omega_0$ directly or just say "in a range", or "in a small range", or something like this}).
While this ensures a bandlimit at the start of training, as it progresses, the layer composition introduces higher frequencies, leading to noisy reconstructions.
We avoid this by providing a novel initialization for the first layer coupled with a bounding scheme which gives controls for limiting the MLP spectrum.
% (Fig~\ref{fig: bounding}) \andre{I would remove the fig reference here, it's quite far and not needed at this point, this is just the RW section}.

Other works \cite{lindell2022bacon, dou2023multiplicative} employed MFNs~\cite{fathony2020multiplicative} to control the bandlimit by applying a filter on the spectrum. However, this strategy usually introduces reconstruction artifacts since MFNs hard-truncate the spectrum.
BANF~\cite{shabanov2024banf} employed a grid-based MLP with a spatial filter that leverages grid resolution to constrain the highest frequency in the network.
In our experiments, we observe that this is prone to creating higher frequencies which propagate as artifacts. 
In contrast, TUNER, grounded in a theoretical result (Thrm~\ref{thm: bound}), guarantees a bandlimited MLP, and serves as a soft filter, providing a representation without ringing artifacts.

%---------------------------------------------------------------------------------
\section{Sinusoidal MLPs as Fourier series}
\label{sec: Theory}
% \andre{Consider rewriting this section as "sinusoidal MLPs" (instead of INRs), since the result is more general than specific INR applications}

This work addresses the problem of deriving an efficient training scheme with controlled bandlimit for sinusoidal MLPs. This section presents the mathematical definitions and novel formulas for approaching this task.

Sinusoidal MLPs demonstrated high representational capacity with only a few hidden layers \cite{sitzmann2020implicit, zell2022seeing, saragadam2022miner}.~To understand how layer composition encodes this capacity, we propose to thoroughly investigate the structure of a {$3$-layer} sinusoidal MLP $f\!:\!\R^{d}\!\to\!\R$. 
% For simplicity, we assume the codomain to be $\R$, however the same analysis holds for dimension $>\!\!1$.
For simplicity, we assume the codomain to be $\R$, however the same analysis holds for dimension $>\!\!1$.
More precisely, we consider $f(\textbf{x})\!=\! \textbf{C}\!\circ\!\textbf{S}\!\circ\!\textbf{D}(\textbf{x})\!+\!e,$ with 
$\textbf{D}(\textbf{x})\!=\!\sin(\omega\textbf{x}\!+\!\varphi)$ being the input layer that projects $\textbf{x}$ into a list of harmonics (input neurons) ${D}_i(\textbf{x})\!=\!\sin\left(\omega_i \textbf{x}+\varphi_i\right)$ with frequencies $\omega\!=\!(\omega_1,\ldots, \omega_{m})\!\in\! \R^{m\times d}$ and shifts $\varphi\!\in\!\R^{m}$.
% \andre{Here, I would suggest explaining in order from right to left, ie from signal input to output, D then S then L. I got confused a little to track that everything makes sense since one needs to go back and forth with the current explanation}
Layer $\textbf{D}$ is then composed with $\textbf{S}(\textbf{x})\!=\!\sin(\textbf{W}\textbf{x} + \textbf{b})$, where $\textbf{W} \in \R^{n\times m}$ is the hidden matrix, and $\textbf{b}\! \in\! \R^{n}$ the bias.
Finally, $\textbf{C}\cdot\textbf{x}+e$ is an affine transformation with $\textbf{C}\in \R^{n\times 1}$ and $e\in\R$.
% We show empirically that such architecture has enough capacity for our purposes.

We now present some properties of the sinusoidal layers, which play key roles in the representation, and give a reinterpretation of them in terms of the network {parameters}.
First, note that the hidden neurons $\textbf{h}(\textbf{x}) := \textbf{S}\circ\textbf{D}(\textbf{x})$ are defined by composing the dictionary $\textbf{D}$ with the hidden sinusoidal layer which results in a list with elements 

\vspace{-0.6cm}

\begin{align}\label{e-neuron}
    h_i(\textbf{x})=\sin\bigg(\sum_{j=1}^{m} {W}_{ij}\sin\left( \omega_j\textbf{x}+\varphi_j\right)+b_i\bigg).
\end{align}

\vspace{-0.3cm}

The following is a key result of this work, which states that we can linearize a hidden neuron \eqref{e-neuron}  as a sum of sines with frequencies and amplitudes determined by $\omega$ and $\W$.  
\begin{theorem}\label{thm: main}
    Each hidden neuron $h_i$ of a {3-layer} sinusoidal MLP has an amplitude-phase expansion of the form
    
    \vspace{-0.5cm}
    
    \emph{
    \begin{align}\label{eq: sumofsines}
        h_i(\textbf{x}) = 
            \sum_{\textbf{k}\in\Z^{m}} \alpha_{\textbf{k}}\,\sin\big(\beta_{\textbf{k}}\,{\textbf{x}} + \lambda_{\textbf{k}}\big),
    \end{align}}
    
    \vspace{-0.4cm}
    
    \noindent where \emph{$\beta_{\textbf{k}} \!=\! \dot{\textbf{k}}{\omega}$}, \emph{$\lambda_{\textbf{k}}\!=\!\dot{\textbf{k}}{\varphi} \!+\! b_i$}, and  \emph{$\alpha_{\textbf{k}} \!=\! \prod_j\!J_{k_j}(W_{ij})$} is the product of the Bessel functions of the first kind.
\end{theorem}
\noindent The Bessel functions $J_k$ appear in the Fourier series of $\sin\big(a\sin(x)\big)$~\cite[Page 361]{abramowitz1964handbook} and Thrm~\ref{thm: main} generalizes this result.
{See the proof and generalization in the Supp. Mat}.
{\citet{yuce2022structured} present a similar expansion for neurons activated by polynomials, however, they do not derive an amplitude formula. Their expansion for sinusoidal neurons is restricted to a 3-layer MLP {without bias}, a critical parameter for Fourier representation. Ours incorporates bias and generalizes to deep MLPs.}

We now list some consequences of \eqref{eq: sumofsines}.
First, the layer composition introduces a vast number of frequencies $\beta_\textbf{k}$ depending only on $\omega$, and shifts given by the input shift $\varphi$ and the bias $\textbf{b}$.
More precisely, truncating the expansion by summing over $\norm{\textbf{k}}_\infty\!\!\leq\!\! B\!\in\! \N$,~\footnote{$\norm{\textbf{k}}_\infty$ denotes $\max\{\norm{k_1}, \ldots, \norm{k_m}\}$.} implies that each hidden neuron $h_i$ could learn $\frac{(2B+1)^m -1}{2}$ non-null frequencies.
This frequency generation gives a novel explanation of why composing sinusoidal layers greatly increases the network capacity. 
% \hl{In the Supp. Mat., we present a toy experiment to show this.}
% \andre{I think this is missing a more explicit connection to the previous sentence, I think you want to say that the hidden layer increases the capacity of the network (or something like that), and this will lead to better fitting}.
We will explore \eqref{eq: sumofsines} to define a robust initialization for the INR's input neurons in Sec~\ref{s-initialization}.

Secondly, note that the weights $\textbf{W}$ fully determine the amplitude $\alpha_\textbf{k}$ of each harmonic $\sin\!\big( \beta_\textbf{k} \textbf{x}\!+\!\lambda_\textbf{k}\big)$. Thus, to derive our bounding scheme we need to focus only on $\textbf{W}$.
Finally, we can derive a sine-cosine form of~\eqref{eq: sumofsines},
\begin{align}\label{eq: quasefourier}
   h_i(\textbf{x}) \!
   =\!\!\! \sum_{\textbf{k}\in\Z^m}\!\!{A}_{\textbf{k}}\sin(
        \beta_{\textbf{k}}\textbf{x}
    ) \!+\!{B}_{\textbf{k}} \cos(
        \beta_{\textbf{k}}\textbf{x}),
\end{align}

\noindent with ${A}_{\textbf{k}} \!=\! \alpha_\textbf{k}\cos(\lambda_\textbf{k})$ and ${B}_{\textbf{k}} \!=\! \alpha_\textbf{k}\sin(\lambda_\textbf{k})$.
Note that the generated frequencies are independent of the index $i$, thus the hidden neurons share the same harmonics: $\sin\big(\beta_{\textbf{k}}\textbf{x}\big)$, $\cos\big(\beta_{\textbf{k}}\textbf{x}\big)$.
Since different combinations of the input frequencies may correspond to a single frequency, \eqref{eq: quasefourier} isn't (yet) the Fourier transform of the network. In other words, we could have two integer vectors $\textbf{k}, \textbf{l}\in\Z^m$ such that $\beta_{\textbf{k}}=\beta_{\textbf{l}}$. In the Supp. Mat., we show how to aggregate those frequencies to obtain the final Fourier~transform.

To control the MLP’s bandlimit, we need a method to bound the amplitudes of $\beta_{\textbf{k}}$. Precisely, for a given number $B>0$ we would need $\norm{\alpha_\textbf{k}}$ to be small for $\norm{\beta_\textbf{k}}\ge B$.
% \andre{This sentence is not clear to me. Why is this an "issue"? Or do you mean something like a "question", something we would desire to have (in order to have control on the bandlimit of the signal)}.
For this, we use the formula of $\alpha_\textbf{k}$ to determine a bounding:
%, showing that if $\norm{\textbf{W}}_\infty\!\!\!<\!2$ then the coefficients ${A}_{\textbf{k}}$ and ${B}_{\textbf{k}}$ decay exponentially as $\norm{\beta_\textbf{k}}_\infty$ grows.

\begin{theorem}\label{thm: bound}  %CORRECT?
    The magnitude of the amplitudes \emph{$\alpha_\textbf{k}$} in the expansion \eqref{eq: sumofsines} is bounded by
    \emph{$
        \prod_{j=1}^m\Big( \tfrac{|W_{ij}|}{2}\Big)^{|k_j|}\frac{1}{|k_j|!}.
    $}
    % \vspace{-0.2cm}
\end{theorem}
Sec~\ref{chap: Bound} presents a bandlimit control mechanism during training based on Thrm~\ref{thm: bound}.

\begin{figure*}[t!]
     \centering
    \includegraphics[width=0.95\textwidth]{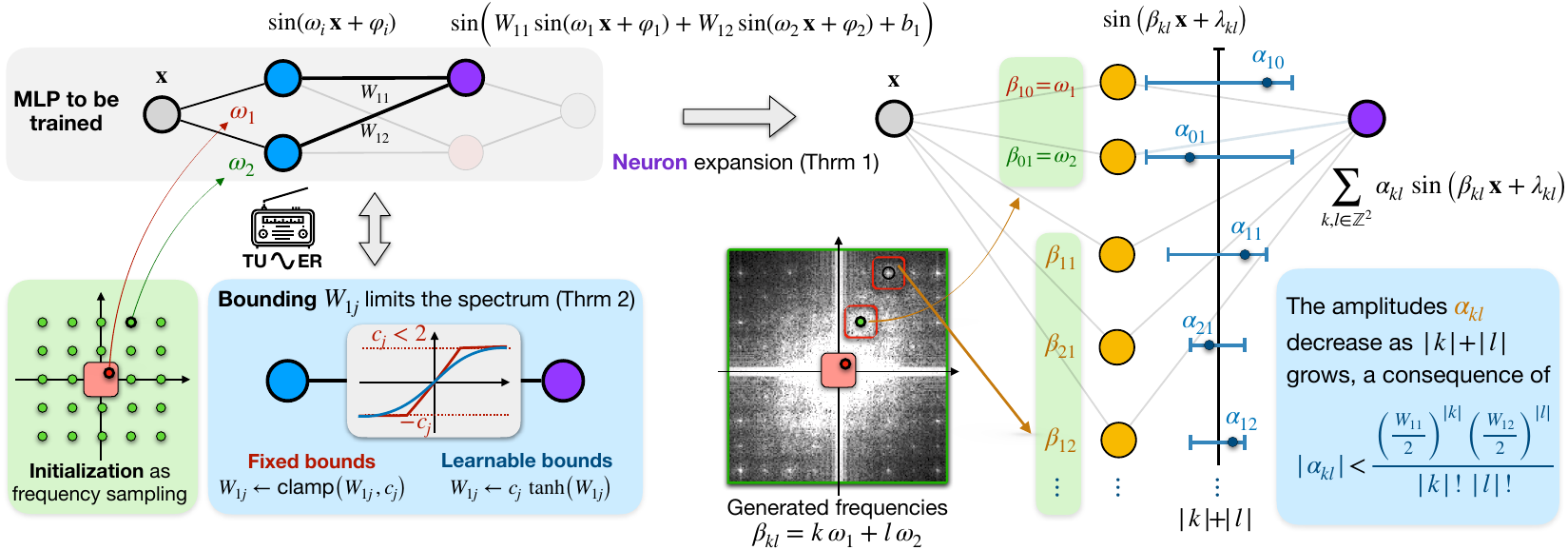}
    \vspace{-0.3cm}
    \caption{\textbf{Overview of TUNER}. 
    % \andre{Small suggestion: put the little radio icon between the red/green arrows and the gray one (ie move it to the left), it would make it a little more clear that both init and bounding are part fo TUNER}
    % \andre{Shouldn't the green dot in the "generated freq" plot refer to beta01 (omega2)? (https://screenshot.googleplex.com/9fzbgFkSHTjWpqd) I don't see why it would refer to beta11}
    % \andre{On the right blue area, maybe replace $c_1$ and $c_2$ by $W_{11}$ and $W_{12}$?}
    % We present TUNER, a robust \textit{training} scheme for sinusoidal MLPs which greatly improves convergence and representation capacity. 
    %
    % TUNER relies on a novel amplitude-phase expansion (Thrm~\ref{thm: main}), which decomposes the MLP into frequencies $\beta_{\textbf{k}}$—integer linear combinations of the input frequencies $\omega_i$ (first layer weights)—and corresponding amplitudes $\alpha_{\textbf{k}}$, fully determined by the hidden weights.
    % We prove that properly bounding the hidden weights $W_{ij}$ filters the MLP since it constrain the amplitudes $\alpha_{\textbf{k}}$. Leveraging this, we enforce a bandlimit on the network during training through two bounding schemes: \textbf{fixed bounds} for cases with known bandlimits, and \textbf{learnable bounds} to optimize the frequency range. The latter, achieved through edge activations, similar to KANs~\cite{liu2024kan}, improves training stability, as our experiments indicate. 
    %
    To train a sinusoidal MLP (gray model, top-left), we employ two techniques derived from Thrms~\ref{thm: main} and \ref{thm: bound}. First, we initialize the input frequencies $\omega$ (green, bottom-left) with a dense distribution of low frequencies (red square) and a sparse distribution of higher frequencies (green grid). 
    This initialization gives flexibility to learn the remaining signal frequencies which are simply integer combinations of $\omega$ (the yellow nodes on the right), a consequence of the amplitude-phase expansion given by Theorem~\ref{thm: main}.
    Note that this initialization resembles a frequency sampling since the training generates those new frequencies around $\omega$.
    Second, we bound the coefficients of the hidden layer weights (blue nodes) to ensure that the MLP remains within a specified bandlimit. This approach is effective because the amplitude-phase expansion (shown on the right) of each hidden neuron (purple nodes) indicates that the amplitudes of the generated frequencies have an upper-bound depending only on the hidden weights (blue, bottom-right).}
    \label{fig: overview}
\end{figure*}

\section{Initialization and frequency bounding}
\label{chap: Init}
We present \textbf{TUNER}, an initialization and frequency control scheme for sinusoidal INRs.
Our initialization considers integer input frequencies, based on a Fourier series interpretation, and uses Thrm~\ref{thm: main} to study the spectrum of layer composition.
Then, we use Thrm~\ref{thm: bound} to control the INR's bandlimit during training.
Fig \ref{fig: overview} provides an overview of TUNER. 
Throughout this section we present toy experiments to motivate the method, using $f\!:\!\R^{2}\!\to\!\R^{3}$ in an image reconstruction setup; later, comprehensive experiments will be given in Sec~\ref{sec: experiments}.
We report the MLP size using the parameters $m$, $n$ as they completely determine the model.
We use the Kodak dataset's images~\cite{kodak} and train the INR using Adam~\cite{kingma2014adam}.
% Additional parameters are described in figure captions.
% \hl{The basic experiment setups are... net (m,n), dataset, adam... \textbf{Diana, pode pegar isso no WandB?}}

\subsection{Initialization as spectral sampling}
\label{s-initialization}

\begin{figure}[t!]%[8]{r}{0.45\textwidth}
     \centering
     \vspace{-0.3cm}
     \begin{subfigure}[b]{0.35\columnwidth}
         \centering
         \includegraphics[width=\textwidth]{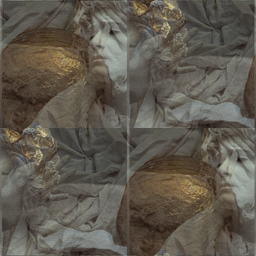}
         % \caption{Using only Steps~\ref{init: step2}, \ref{init: step4}.}
     \end{subfigure}
          \begin{subfigure}[b]{0.35\columnwidth}
         \centering
         \includegraphics[width=\textwidth]{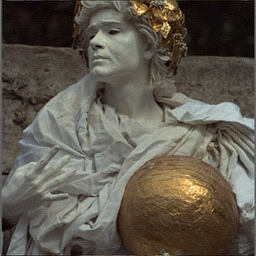}
         % \caption{Using the freqs in (a).}
     \end{subfigure}
     \centering
     \vspace{-0.3cm}
    \caption{Choosing $\omega$ as the cartesian product of the odd frequencies without (left) or with (right) the frequencies $(1,0), (0,1)$. Note that adding them prevents sub-periods (see Supp. Mat.). We trained for $3000$ epochs on a $256^2$ image with network parameters $m=72$, $n=512$, and $\gt{b}=30$.
}
\vspace{-0.2cm}
    \label{fig: subperiods_init}
\end{figure}

\begin{figure}[t!]
     \centering
     \vspace{-0.2cm}
\includegraphics[width=0.95\columnwidth]{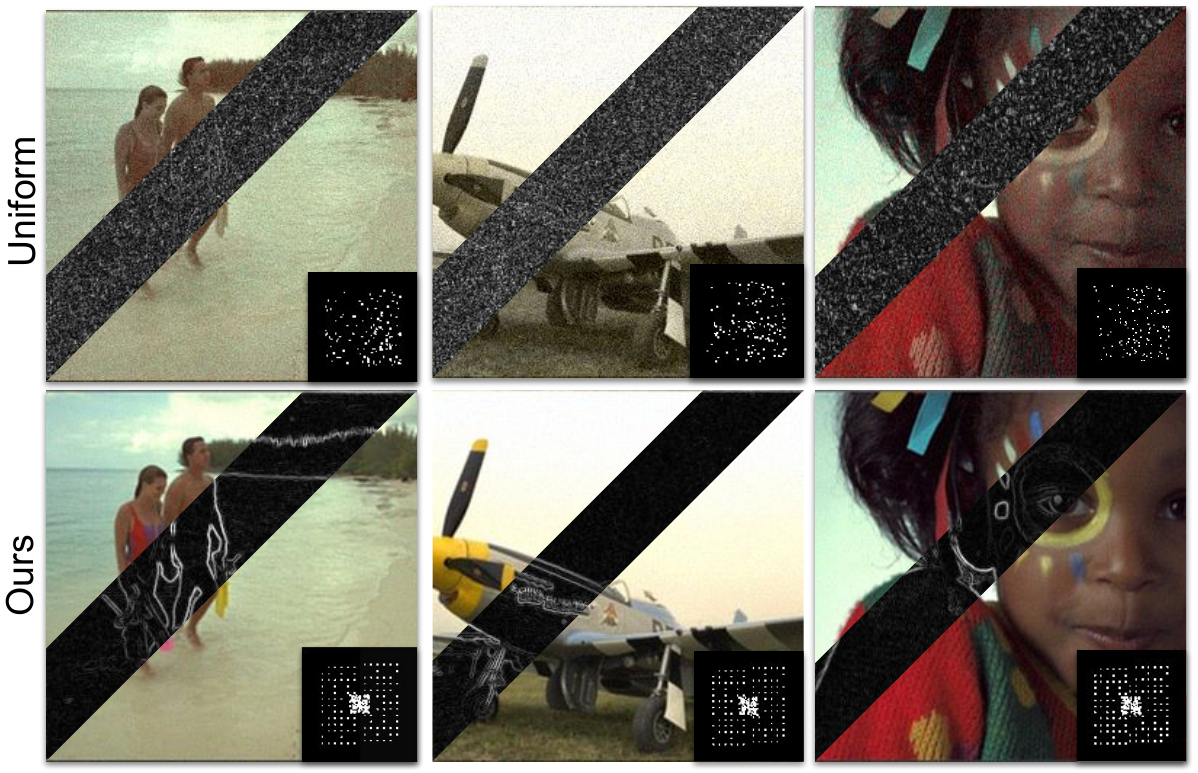}
          \vspace{-0.3cm}
         \caption{{Uniform init. of $\omega$ (top) and ours (bottom). Grayscale bands show INR's gradient.} Ours offers better signal/gradient reconstruction w/o gradient supervision. The MLPs $(m\!=\!128, n\!=\!256)$ with bandlimit $\gt{b}\!=\!87$ were trained for $3000$~epochs.}
         \label{fig: comparison_unif_ours}
         \vspace{-0.6cm}
\end{figure}

A key challenge in initializing a sinusoidal MLP lies in defining its input frequencies $\omega$. Recall that the MLP generates frequencies as integer combinations of $\omega$, i.e., $\beta_{\textbf{k}} \!=\! \sum_i k_i \omega_i$, with the remaining weights representing their amplitudes $\alpha_{\textbf{k}}$. This setup presents two main challenges. First, the initialization of $\omega$ must enable $\beta_{\textbf{k}}$ to cover the signal’s spectrum; otherwise, the optimization cannot learn missing frequencies. Fig~\ref{fig: subperiods_init} (left) shows an example where using only odd frequencies leads to poor reconstruction.
The second problem is the lack of flexibility for learning new frequencies. For example, randomly initializing $\omega$ may not be enough since it may produce many high frequencies in order to overfit the signal, see Fig~\ref{fig: comparison_unif_ours}~(top).

% , while our initialization (right) enables representation of the entire spectrum. 
% \andre{I think this is too early to talk about our method, as we haven't presented it yet. We can keep talking about the challenges only and illustrating them with images, and come back to refer to the ways we resolve them later}.

% To ensure that all frequencies can be flexibly represented we split the sample in two part, a dense sampling of low frequencies a sparse sample of higher frequencies, see Fig~\ref{fig: comparison_unif_ours}(right).

% We detail the initialization scheme \andre{We just talked about the challenges before. Instead of starting with "We detail the initialization scheme", we can say something like "To address these challenges, we introduce a new initialization scheme" -- this "glues" better the story and connects what we introduce to the challenges we just discussed}. 
To address these challenges, we introduce a novel initialization.
First, we restrict our MLP to periodic functions, ensuring the training domain is in a full period. Such functions can be represented by sums of sines with integer frequencies, defining an orthogonal basis (Fourier series); 
This~aligns well with our analysis (Thrm~\ref{thm: main}). The main challenge is initializing $\omega$ such that the generated frequencies $\beta_{\textbf{k}}$ cover the full spectrum within a bandlimit $B$ (Fourier~basis). 
% Then, optimization will learn the corresponding amplitudes (Fourier coefficients) \andre{maybe this sentence can be removed, you already mention this right after eq 4}.

More precisely, we initialize the weights $\omega$ with integer frequencies, that is $\omega_j \!\!\in\!\! \frac{2\pi}{p}\Z^{d}$,
and freeze them during training, with $p\!>\!0$ being the period of the INR.
This guarantees that the input neurons are $p$-periodic and Thrm~\ref{thm: main} ensures that such periodicity is preserved over layer composition. 
Additionally, \eqref{eq: quasefourier} implies that we can rewrite the INR as:
\begin{align}\label{eq: inr_fourier}
   f(\textbf{x})\! =\!\!\sum_{\textbf{k}\in\Z^m}\!\!\dotprod{\textbf{C},{A}_{\textbf{k}}}\sin\big(\beta_{\textbf{k}}\textbf{x}\big)\! +\! \dotprod{\textbf{C}, {B}_{\textbf{k}}} \cos\big(\beta_{\textbf{k}}\textbf{x}\big)\! +\! e.
\end{align}
\noindent Since the generated frequencies $\beta_{\textbf{k}}$ only depend on the frozen parameters $\omega$, the training is responsible for learning the amplitudes of the sine-cosine series in~\eqref{eq: inr_fourier}.

Let $B$ be the signal's bandlimit. To ensure that the generated frequencies do not bypass $B$, we cannot sample $\omega$ directly on $[-B, B]^d$ since the MLP generates multiples of these frequencies. 
Therefore, we sample $\omega_{j}$ in $\frac{2\pi}{p}[-\gt{b}, \gt{b}]^d$, with $\gt{b}\in(0,B)$ being a threshold for the input frequencies.
Indeed, as the coefficients of hidden matrix $\W$ are limited by $2$ (see Sec \ref{chap: Bound}), the~magnitudes of ${A}_\textbf{k}$ and ${B}_\textbf{k}$ decrease as $\norm{\textbf{k}}_\infty$ increases.
This~is a consequence of Thrm~\ref{thm: bound}:
\vspace{-0.1cm}
\begin{align}\label{eq: coef_decay}  % CORRECT?
    |\alpha_{\textbf{k}}| \leq \prod_{j=1}^m\!\frac{\Big( \frac{|W_{ij}|}{2}\Big)}{|k_j|!}^{|k_j|}\!\!\!\!\! \leq \!\frac{1}{|k_1|!\ldots|k_m|!}; \,\, i=1,\ldots,n.
\vspace{-0.1cm}
\end{align}
Thus, the generated frequencies $\beta_{\textbf{k}}$ with small $\textbf{k}$ have more influence over the expansion \eqref{eq: inr_fourier}, mimicking  Fourier~series.

\begin{figure}[!t]
     \centering
     \includegraphics[width=\columnwidth]{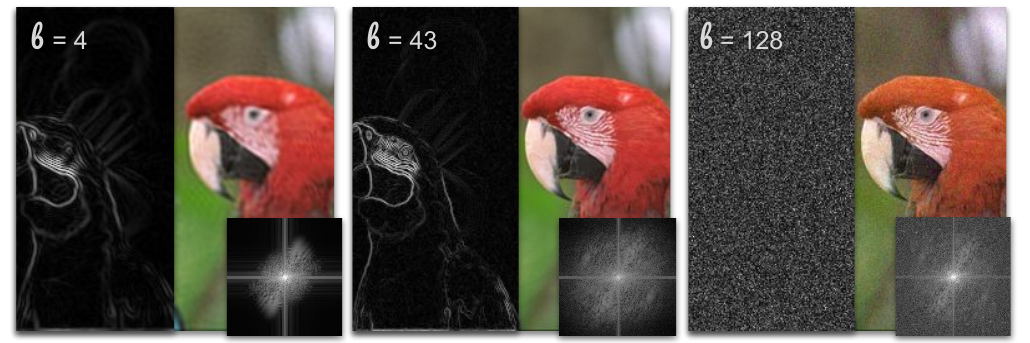}
      \vspace{-0.6cm}
    \caption{Image reconstructions with bandlimit $B\!\!=\!\!128$, varying $\gt{b}\!=\!4,43,128$ with a network ($m\!=\!80, n\!=\!1000$) trained over 3000 epochs. 
    The gradient magnitude is shown on the left of each image. Note that smaller $\gt{b}$ (middle-left) yield better reconstructions, while higher values of $\gt{b}$ introduce noisy gradients (right).}
   \vspace{-0.5cm}
    \label{fig: init_bandlimit}
\end{figure}

\noindent For example, if input frequencies include 1 and 20, frequencies like 19 = 20 - 1 and 21 = 20 + 1 can be represented through their combinations.
Thus, low frequencies give flexibility in expanding around the input ones which motivates initializing more input frequencies near the origin.
We found $\gt{b}=\frac{B}{3}$ to work well experimentally.
Fig~\ref{fig: init_bandlimit} shows that it is necessary, as higher $\gt{b}$ makes it harder to learn the small frequencies, introducing noise.
% In the next section we avoid this problem by bounding the INR's spectrum.

To initialize $\omega\!\in\!\R^{m\times d}$, we need to choose $m$ frequencies in $[-\gt{b}, \gt{b}]^d$. Taking into account the importance of the lower frequencies discussed above, we split it into regions $\textbf{L}\!=\![-\gt{l}, \gt{l}]^d$ of low and $\textbf{H}=[-\gt{b}, \gt{b}]^d\setminus\textbf{L}$ of high frequencies, $\gt{l}\in(0,\gt{b})$.
% delimited by $0<\gt{l}<\gt{b}$ (Fig~\ref{fig: spectrum_partition}). 
% Then, given the percentages $p_1, p_2$ ($p_1+ p_2=1$, $p_i\geq 0$), we sample $m$ points in $\textbf{K}$ with $\lceil p_1m\rceil$ low and $\lfloor p_2m\rfloor$ high input frequencies.
As $\beta_\textbf{k}$ only has a significant amplitude when
$|\textbf{k}|_\infty$ is small  \eqref{eq: coef_decay}, initializing with only low input frequencies may not allow the generation of high frequencies. Thus, we initialize some frequencies in $\textbf{H}$ to cover the spectrum, and the rest in $\textbf{L}$ to cover a neighborhood of the input frequencies.
Specifically, we select $m_l<m$ frequencies uniformly from $\textbf{L}$. Then, we choose $m-m_l$ frequencies from $\textbf{H}$, spread in a grid-like fashion to cover $[-\gt{b}, \gt{b}]^d$. In our experiments we considered $m_l=0.7 \,m$.
Finally, since $\textbf{D}(\x)\!=\!\sin(\omega\x\!+\!\varphi)$, we initialize $\varphi$ uniformly in $[-\frac{\pi}{2}, \frac{\pi}{2}]$.
% Adding $(1,0)$ and $(0,1)$ to the input frequencies has a huge impact since it allows us covering the whole spectrum resulting in better reconstruction, see Fig~\ref{fig: subperiods_init}(right).

\begin{figure}[h!]%[11]{r}{0.75\textwidth}
         \centering
         \vspace{-0.2cm}
         \includegraphics[width=\columnwidth]{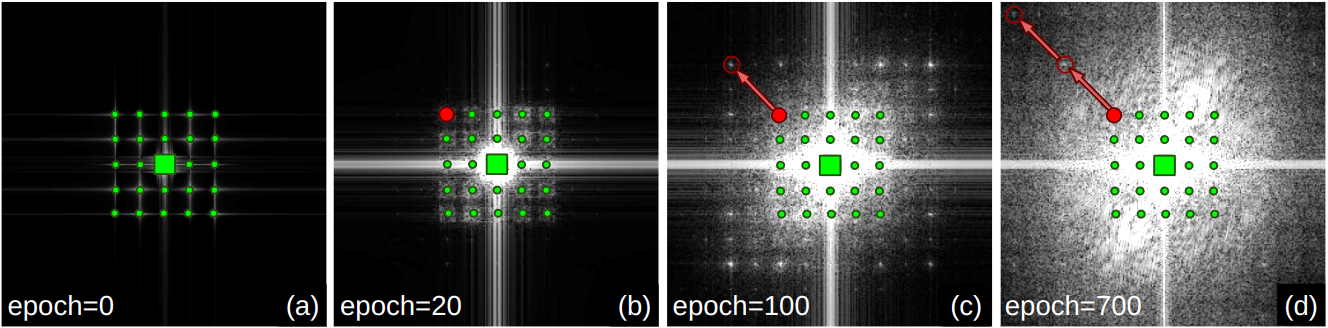}
         \vspace{-0.5cm}
        \caption{INR spectrum during training. The input frequencies $\omega$ in green. During training new frequencies appear in a neighborhood of $\omega$. The red dot indicates a high input frequency $\omega_j$ and the red arrows show the generated frequencies $k_j\omega_j$, $k_j=1, 2, 3$.}
    \label{fig: init_exponential}
    \vspace{-0.1cm}
\end{figure}

When $\omega_j\!\in\!\textbf{L}$, small variations in $k_j$ results in a frequency near $\beta_\textbf{k}$. Indeed, Fig~\ref{fig: init_exponential} shows how low frequencies (green square) introduce frequencies around the input frequencies (Fig~\ref{fig: init_exponential}a).
Conversely, if $\omega_j\in\textbf{H}$ then $\beta_{\textbf{k}+\textbf{e}_j}$ is far away from $\beta_\textbf{k}$; we note this in Fig~\ref{fig: init_exponential}b-d, where $\omega_j$ is the red point and the arrows show the generated frequencies $k_j\omega_j$.

Finally, when sampling frequencies in \textbf{L}, we include $(1,0)$ and $(0,1)$ to guarantee that any frequency could be generated and prevents the appearance of sub-periods as shown at the beginning of this section in Fig~\ref{fig: subperiods_init} (left) (see also Supp. Mat for more details).
Fig~\ref{fig: subperiods_init} (right) shows how this problem is solved using our initialization.

The proposed initialization overcomes the noisy reconstruction issues that arise with naive uniform sampling in $[-\gt{b}, \gt{b}]^d$, as used in \cite{sitzmann2020implicit} (Fig~\ref{fig: comparison_unif_ours} (top)). Fig~\ref{fig: comparison_unif_ours} (bottom) demonstrates that our scheme significantly improves both signal and gradient reconstruction quality.

\subsection{Bounding and hidden layer initialization}
\label{chap: Bound}

\begin{figure}[!h]
% \vspace{-0.2cm}
     \centering
         \includegraphics[width=0.85\columnwidth]{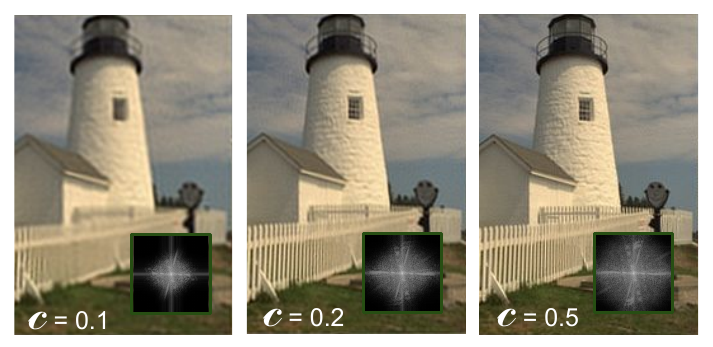}
     \vspace{-0.2cm}
    \caption{Training a network $f$ with $m=n=256$ for $6000$ epochs with $\gt{b}\!\!=\!\!30$ and bounds $\gt{c}\!\!=\!\!0.1, 0.2, 0.5$. The bound over the hidden matrix $\W$ restricts the appearance of high frequencies.}
    \label{fig: bounding}
    \vspace{-0.2cm}
\end{figure}

As shown in previous sections, sinusoidal INRs can represent detailed signals due to the wide number of frequencies generated by layer composition.
However, they could manifest as noise in the reconstruction, see Fig \ref{fig: init_bandlimit} (right).
% \andre{are you pointing to the correct figure here? Is the appearance of new high freqs in Fig 6 necessarily noise?}{Corrected the figure reference}.
Specifically, high input frequencies may generate even higher ones, bypassing the bandlimit $B$.
To avoid this, we use Thrm~\ref{thm: bound} to introduce a bounding 
scheme to limit the amplitudes.
Precisely, Thrm~\ref{thm: bound} implies that in each hidden neuron $h_i$, the amplitude of a generated frequency $\beta_\textbf{k}$ is limited by:
\begin{equation}\label{eq: norm_inf_bound}
    |\alpha_\textbf{k}|\leq \left(\dfrac{\norm{\W_i}_\infty}{2}\right)^{\norm{\textbf{k}}_1}\frac{1}{\prod_{j=1}^m|{k_j}|!}.
\end{equation}
where $\W_i$ is the $i$-row of $\W$.
Note that $\norm{\W}_\infty\!\leq\!2$ implies that the amplitude of any generated frequency $\beta_\textbf{k}$ rapidly decreases when the coordinates of $\textbf{k}$ grow.
Therefore, we use \eqref{eq: norm_inf_bound} to define a bounding scheme that only allows the appearance of generated frequencies with small multiples of~$\omega$ which is beneficial as shown in previous sections. 
For this, we simply clamp the coefficients $W_{ij}$ by a bound parameter $\gt{c}\in(0,2]$ at each optimization step, enforcing $\norm{\W}_{\infty}\leq\gt{c}$.
% Then,~again, \eqref{eq: norm_inf_bound} implies that high order frequencies have very small amplitudes.
Fig~\ref{fig: bounding} shows results of bounding $\norm{\W}$ with different~$\gt{c}$.

% We initialize $\W$ with values in $\left[-\gt{c}, \gt{c}\right]$ using a normal distribution with mean 0 and standard deviation~$\frac{\gt{c}}{3}$.
% Since SIREN~\cite{sitzmann2020implicit} initializes $\W$ uniformly in $\big(\!\!-\!{\sqrt{{6}/{m}}}, {\sqrt{{6}/{m}}}\big)$, and as $\norm{\W}_\infty\!\leq\!\sqrt{{6}/{m}}<2$ for $m>1$, it is bandlimited only at the beginning of training.
% Conversely, our initialization controls the bandlimit during all training and has faster convergence (see Supp. Mat.).
% Observe that 
% SIREN \cite{sitzmann2020implicit} also controls the bandlimit but only at initialization, since the weights of $\W$ are sampled in $\big(-\sqrt{6/m}, \sqrt{{6}/{m}}\big)$. 
% In other words, since $\norm{\W}_\infty<2$ (for $m>1$) at initialization, they do not introduce high frequencies at the beginning of the training. Additionally, we note a faster convergence when training with bounding using our initialization (see Supp. Mat.).

% This bounding is straightforward, but does not consider the roles that different frequencies play during reconstruction. 
% In Sec~\ref{s-initialization} we saw that the lower frequencies initialized in the set $\textbf{L}$ have a main role in the frequency generation.
% Then, we use the relationship between the hidden matrix $\W$, generated frequencies $\beta_\textbf{k}$, and $\omega$ to define the bounding by blocks.
In Sec~\ref{s-initialization} we split the input frequencies $\omega$ in two sets: the low ($\norm{\omega_j}_\infty\!\leq\!\gt{l}$) and high ($\norm{\omega_j}_\infty\!>\!\gt{l}$) frequencies.
Additionally, we saw that the lower ones have a main role in the frequency generation, thus, we consider a higher threshold for them. 
% Thus, we propose to consider different bounding thresholds: higher 
For this, observe that each input frequency $\omega_j$ is associated with the $j$-column 
 of $\W$.
Then, bounding the columns of $\W$ related to high frequencies, we restrict the amplitudes of $\beta_\textbf{k}$ that could exceed the Nyquist limit $B$. Conversely, since low frequencies are used to span around~$\omega$, we consider a higher bound for their columns.
Precisely, we define two parameters $\gt{c}_{\textbf{L}}$ and $\gt{c}_{\textbf{H}}$ to bound the weights $W_{ij}$ corresponding to $\omega_j$ in \textbf{L} and \textbf{H}, respectively,

\vspace{-0.6cm}
\begin{align*}
\small
    &&\norm{\omega_j}_{\infty}\!\leq\!\gt{l} \,\,\longrightarrow\,\, W_{ij}= \texttt{clamp}\big(W_{ij}, [-\gt{c}_{\textbf{L}},\gt{c}_{\textbf{L}}]\big),\\
    &&\norm{\omega_j}_{\infty}\!>\!\gt{l} \,\,\longrightarrow\,\, W_{ij}= \texttt{clamp}(W_{ij}, [-\gt{c}_{\textbf{H}},\gt{c}_{\textbf{H}}]\big).
\end{align*}

\vspace{-0.2cm}

We use these bounds to initialize $\W$ such that each column is normal distributed with mean $0$ and standard deviation $\tfrac{\gt{c}_{*}}{3}$, with $\gt{c}_{*}$ being the bound corresponding to that column.
Note that SIREN~\cite{sitzmann2020implicit} initializes $\W\in\R^{n\times m}$ uniformly in $\small{\big(\!\!-\!{\sqrt{{6}/{m}}}, {\sqrt{{6}/{m}}}\big)}$, and as $\small{\norm{\W}_\infty\!\leq\!\sqrt{{6}/{m}}<2}$ for $m>1$, it is bandlimited only at the beginning of training, since $\W$ is not controlled in any way during the optimization.
Conversely, our initialization controls the bandlimit during all training and has faster convergence.% (see Supp. Mat.).

Additionally, we present a scheme to learn the bounding parameters during training. For this, we bound each $j$-column of the hidden matrix $\textbf{W}$ using a $\tanh$ activation followed by multiplication with a trainable bounding parameter $\gt{c}_j$. Clearly, the resulting column is bounded by~$\gt{c}_j$. In other words, we replace the hidden layer 
$\sin(\textbf{W}\textbf{x} + \textbf{b})$ by

\vspace{-0.6cm}

\begin{align}\label{e-learnable-boundings}
\small
    \sin\big(\tanh(\textbf{W})\gt{C}\textbf{x} + \textbf{b}\big),
\end{align}

\vspace{-0.2cm}

\noindent where $\gt{C}$ is a diagonal matrix with each $jj$-entry being~$\gt{c}_j$.
To prevent the bounds from growing too large, we use Thrm~\ref{thm: bound} to define the regularization term as $\mathcal{L}_{\textrm{reg}} = \sum |\gt{c}_j|$.
Otherwise, the hidden weights may grow unbounded, leading to the INR's overfit as in the usual training.

\section{Experiments}
\label{sec: experiments}

% \andre{Suggestion: rename this section to simply "Experiments"}

This section presents experiments regarding the initialization and bounding of frequencies of an INR~$f$ considering the images from Kodak dataset \cite{kodak} resized to $[-1,1]^2$. 
We describe the architecture of $f$ with hidden matrix $\W\!\in\!\R^{n\times m}$ with the parameters $m,n$. 
% Since our architecture is determined by the hidden matrix $\W\in\R^{n\times m}$, we use $n$ and $m$ to define $f$. 
{We initialize $f$ following the scheme in Sec~\ref{s-initialization} and optimize it using Adam~\cite{kingma2014adam} with learning rate $10^{-4}$.}
For the quantitative evaluations, we report the mean and standard deviation of the Peak Signal-to-Noise Ratio (PSNR) evaluated on the test set ($10\%$ random pixels).
{To show that TUNER adds high order regularity to the reconstruction, we compare the analytical gradient $\nabla f$ to the Sobel filter of the ground truth. All experiments were run in a 12 GB NVIDIA GPU (TITAN X Pascal)}.

% \vspace{-0.3cm}

\subsection{Initialization}
\label{chap: experiments_init}
First, we compare our initialization of $\omega$ {($36.2$dB)} against the uniform distribution {($32.2$dB)}.
We obtain a $4$dB improvement of PNSR on average (with a std of $1.6$dB). 
We also present an ablation over the parameter $\gt{l}$ which splits the square $[-\gt{b}, \gt{b}]^ 2$ in regions of low ($\textbf{L}$) and high ($\textbf{H}$) frequencies.
Here we use $\gt{b}=85$ and $\gt{l}\!=\!21$, $42$, $64$, 
% an INR with $m\!=\!52$, $n\!\!=\!\!256$, 
and sample $\omega$ with $30\%$, $50\%$, and $70\%$ of coordinates in \textbf{H}.
Table~\ref{table: low_freqs_limit} shows the performance of the INR.
% , with the signal (grad.) mean std of $1.9$dB ($3.1$dB).

\begin{table}[ht!]
% \begin{wraptable}[10]{r}{0.5\textwidth}
% \scriptsize
% \setlength{\tabcolsep}{4pt}
    \centering
    \footnotesize
    \begin{tabular}{c|cc|cc|cc}
    \hline
    \multicolumn{1}{l|}{{$\%$ high freqs}} & \multicolumn{2}{c|}{$30\%$} & \multicolumn{2}{c|}{$50\%$} & \multicolumn{2}{c}{$70\%$}                            \\ \cline{1-7} $\gt{l}$
                          & Signal        & Grad        & Signal        & Grad        & \multicolumn{1}{c}{Signal} & \multicolumn{1}{c}{Grad} \\ \hline
    \hline
    $21$ & \textbf{35.2} & \textbf{26.9} & \textbf{35.2} & \textbf{26.6} & \textbf{35.0} & \textbf{26.0} \\ \hline
    $42$ & 35.1 & 26.3 & 34.8 & 24.6 & 33.8 & 22.2 \\ \hline
    $64$ & 34.2 & 25.2 & 33.3 & 21.6 & 30.8 & 15.0 \\ \hline
    \end{tabular}%
    \vspace{-0.2cm}
    \caption{Training an INR with $m=n=416$ parameters, for $400$ epochs. 
    Results indicate that it is better to sample more $\omega_j\in\textbf{L}$ and choose a smaller $\gt{l}$.
    % {Add m, n}}
    }
    \label{table: low_freqs_limit}
    \vspace{-0.5cm}
\end{table}

This experiment considered all the images in the dataset.
Observe that all values of the row $\gt{l}\!=\!64$ have lower PSNR, showing that an INR with very high frequencies performs worse even for small percentages ($30\%$). 
Similarly, the performance worsens when we increase the number of high frequencies (last column).

\begin{figure}[h!]%[7]{r}{0.5\textwidth}
     \centering
         \centering
         \includegraphics[width=0.8\columnwidth]{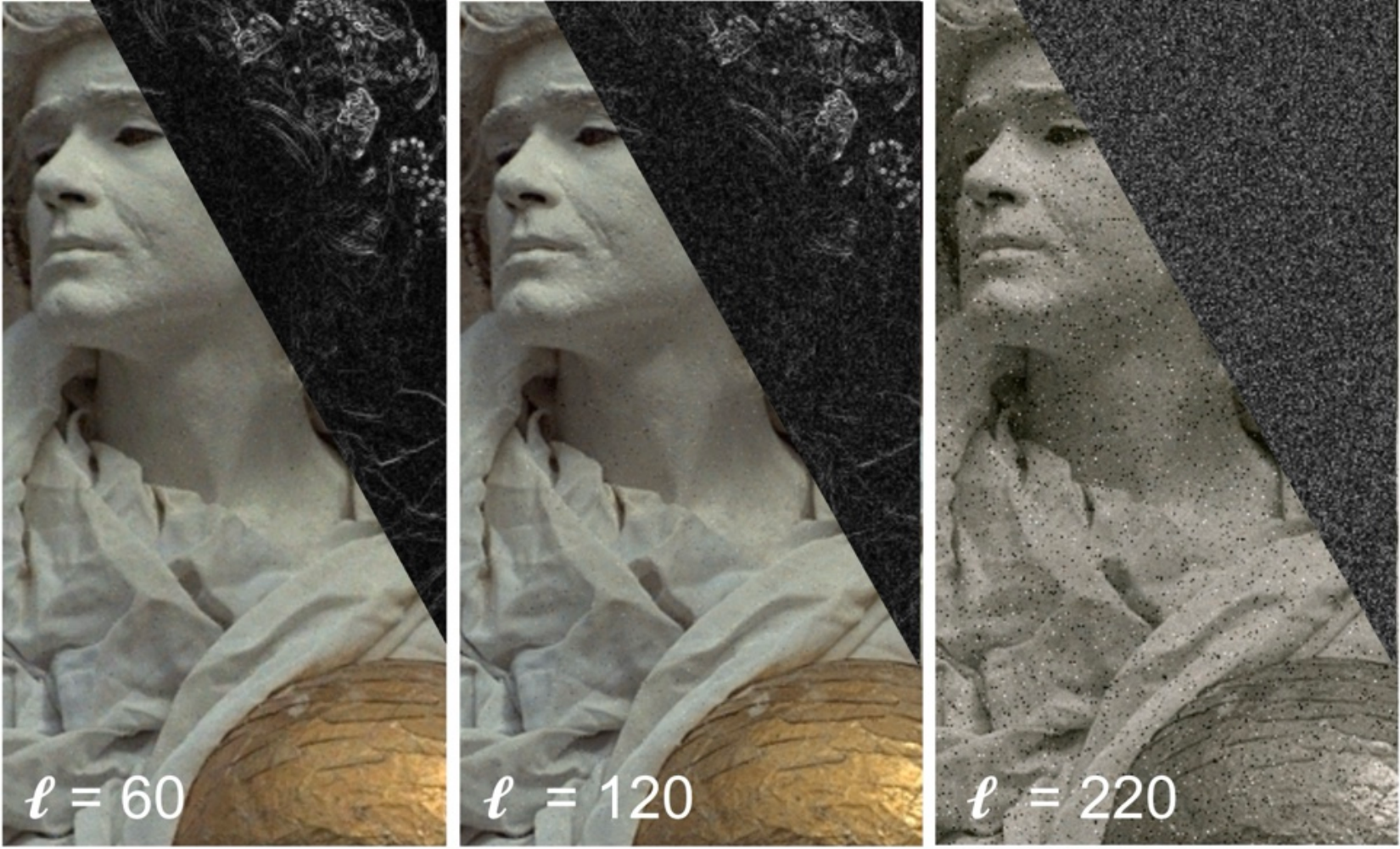}
    \vspace{-0.2cm}
     \caption{Reconstructions with big $\gt{l}$ present overfitting on the $10\%$ unsupervised points.
     {Gradient on the upper right corner.}
     % We train the INRs on $90\%$ of the pixels.
    }
    \label{fig: low-high}
    \vspace{-0.4cm}
\end{figure}

We now test the influence of high frequencies by training an INR of size $m\!=\!360$, $n\!=\!512$ with $\gt{b}\!=\!256$ over $90\%$ of the ground truth pixels. We vary the parameter $\gt{l}\!=\!60, 120, 220$ that defines low frequencies.
% , that split the frequencies into low and high.
% and bound these with values $\gt{c}_{\textbf{L}}=1.4, \gt{c}_{\textbf{H}}=0.1$.
Increasing the number of higher frequencies, the INR suffers from noise and overfitting. This can be noted in the reconstruction~on the unsupervised pixels and gradients in~Fig~\ref{fig: low-high}.
Finally, in Supp. Mat. we compare our initialization of $\W$ against SIREN and observe a faster convergence.

\subsection{Bounding}

% For each threshold, we initialize $\omega$ as the Cartesian product of a list of ten $1D$ frequencies. We consider $l$ of the ten frequencies to be high frequencies sampled randomly from the interval $[\gt{l}, \gt{b}]$, with $l=2, 4, 6, 8$.

In Sec~\ref{chap: Bound} we presented a scheme to bound the INR spectrum.
Here, we show additional experiments to demonstrate that this improves the signal and has a great benefit over the unsupervised gradient reconstruction.
Then, we give a comparison between fixed and trained bounding schemes.

\begin{figure}[!h]
% \begin{wrapfigure}[12]{r}{0.5\textwidth}
         \centering
        %  \vspace{-0.3cm}
         \includegraphics[width=0.9\columnwidth]{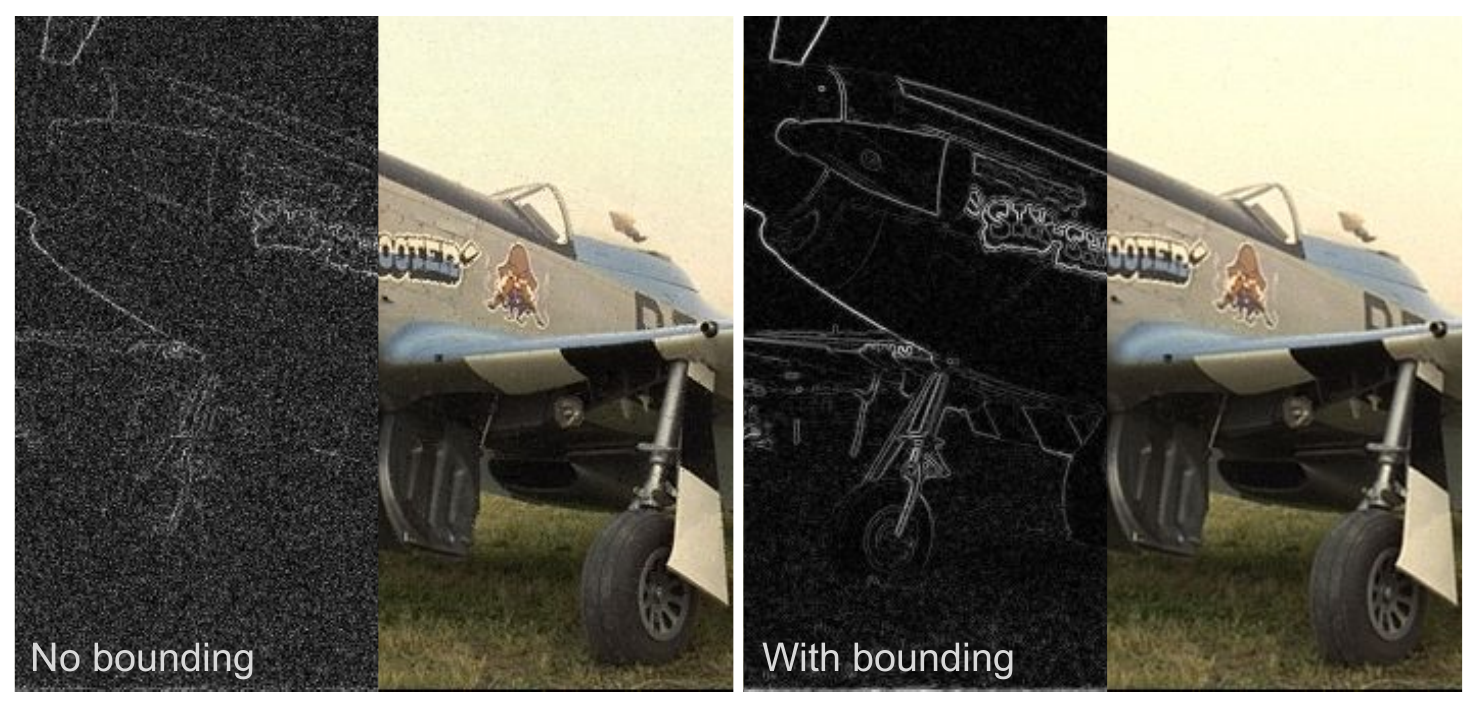}
    % \centering
    \vspace{-0.3cm}
    \caption{Comparison of training wo/w {fixed} bounds. Note that the use of bounds preserves high order information of the signal. {Left side of the figures show the network gradients.}
    }
    % \label{table: bounds_ablation}
    \label{fig: bounds}
\vspace{-0.3cm}
\end{figure}

Fig~\ref{fig: bounds} compares the reconstructions of an image of size $512^2$ wo/w our {fixed} bounding scheme. To demonstrate the impact of our scheme, we sample the input frequencies using $\gt{b}\!=\!256$ and $\gt{l}\!=\!100$. 
Without bounding (left), we effectively reconstruct the signal at the supervised pixels (30.8 dB), but with a noisy gradient (15.8~dB). That is, we are overfitting the signal. 
Conversely, bounding the hidden weights significantly improves both the signal (35.1 dB) and its gradient (27.8~dB).
Our bounding scheme acts like a filter in both signal and gradient reconstruction.

We compare quantitatively our bounding scheme over the dataset, varying the parameters
 $\gt{b}$ and $\gt{c}_{\textbf{H}}$. 
Table \ref{tab: bounds_ablation} shows an improvement in the signal for all cases. 
Furthermore, the unsupervised gradient of the image is better learned when using {fixed} bounds, showing an average increase of 9.6dB against the standard training.
Also, the last two lines of the table show the impact of the parameters $\gt{c}_{\textbf{L}}$, $\gt{c}_{\textbf{H}}$ over the signal and gradient reconstructions. 
We consider training with $\gt{c}_{\textbf{L}}>\gt{c}_{\textbf{H}}$ since we are prioritizing low input frequencies in the spectrum generation.
Otherwise, this may generate high frequencies, possibly bypassing the Nyquist limit and resulting in a worse gradient reconstruction.
\begin{table}[h!]%[12]{r}{0.5\textwidth}
\footnotesize
\centering
% \vspace{-0.3cm}
    \begin{tabular}{lll|llll}
    \cline{4-7}
                         &                         &    & \multicolumn{2}{c}{signal} & \multicolumn{2}{|c}{grad} \\ \hline 
    \multicolumn{1}{l}{$\gt{b}$} & \multicolumn{1}{l}{$\gt{c}_{\textbf{L}}$} & $\gt{c}_{\textbf{H}}$ & \multicolumn{1}{|l}{wo bound}   & w bound   & \multicolumn{1}{|l}{wo bound}    & w bound   \\ \hline \hline
    \multicolumn{1}{l}{256}  & \multicolumn{1}{l}{1.5}   &  0.05  & \multicolumn{1}{|l}{32.6}      &  34.8    & \multicolumn{1}{|l}{14.5}     &  25.8    \\ \hline
    \multicolumn{1}{l}{190}  & \multicolumn{1}{l}{1.5}   & 0.05   & \multicolumn{1}{|l}{34.0}      &  \textbf{35.9}   & \multicolumn{1}{|l}{18.5}     &   \textbf{26.1}  \\ \hline
    \multicolumn{1}{l}{190}  & \multicolumn{1}{l}{1.5}   & 0.2   & \multicolumn{1}{|l}{34.0 }     &   34.4  & \multicolumn{1}{|l}{18.5}      &  21.0    \\ \hline
    % \multicolumn{1}{l}{190}  & \multicolumn{1}{l}{1.0}   & 0.05   & \multicolumn{1}{l}{34.0 }     &   35.5  & \multicolumn{1}{l}{18.5}      &  26.0    \\ \hline
    \end{tabular}
    \vspace{-0.3cm}
        \caption{Training with bounding improves the signal and grad. reconstruction. The choice of {the fixed bound} $\gt{c}_{\textbf{H}}$ greatly impacts the grad. fidelity. The std of the difference between each image signal (grad) quality wo/w bounding is $1.4$dB (1.6dB) on average. Evaluations performed on the test set ($10\%$ of pixels).}
    \label{tab: bounds_ablation}
\end{table}

% {orange}{Experiments:

% 1. Fixed bounds and learned bounds.

% 2. Test with other loss function term for learned bounds.}

\begin{figure}[h!]%[13]{r}{0.69\textwidth}
% \vspace{-0.3cm}
         \centering
         \includegraphics[width=\columnwidth]{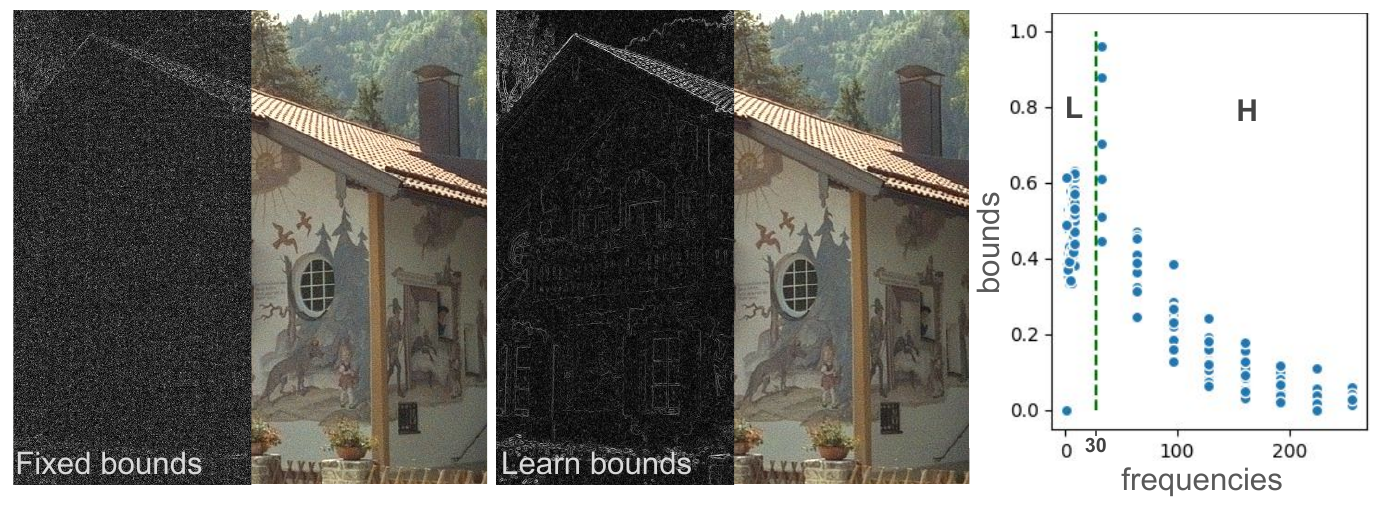}
    % \centering
    \vspace{-0.7cm}
    \caption{Comparison of signal and gradient {(grayscale)} when training with fixed/learned bounds. The learned bounds adjust the INR spectrum resulting in better reconstructions. Each blue point represents a pair {$\left(|\omega_j|_\infty, \gt{c}_j\right)$, where $c_j$ is the trained bound of $\omega_j$.}}
    % \label{table: bounds_ablation}
    \label{fig: learn_bounds}
\end{figure}

Next, we show how we can learn each bound $\gt{c}_j$. 
As described in Sec \ref{chap: Bound}, we can incorporate a learnable $\gt{c}_j$ for each $j$-column of the hidden matrix through a modified layer \eqref{e-learnable-boundings}. 
Fig~\ref{fig: learn_bounds} shows a comparison between our fixed bound scheme using initial un-tuned $\gt{c}_{\textbf{L}}\!\!=\!\!1.0$ and $\gt{c}_{\textbf{H}}\!\!=\!\!0.6$ (left) and the learned bound scheme (middle). 

As expected, training lowers the bounds for high frequencies ($\textbf{H}$) while maintaining or increasing the bounds of low frequencies~($\textbf{L}$). Thus, we achieve better reconstruction in both the signal and gradient.
In Fig \ref{fig: learn_bounds} (right), we note that low frequencies (left of the green line) have significant bounds, showing their importance during training.
These schemes allow us to initialize with higher frequencies than $\nicefrac{B}{3}$, accelerating convergence.
{Finally, we performed an ablation study on the regularization term $\mathcal{L}_{reg}$ to show its importance. Training w/wo regularization yielded $28.5/27.4$dB for the \texttt{RGB} and $27.1/24.0$dB for the gradient.}

\subsection{Comparisons}

\begin{figure}[h!]
\centering
    \vspace{-0.1cm}
    \includegraphics[width=\columnwidth]{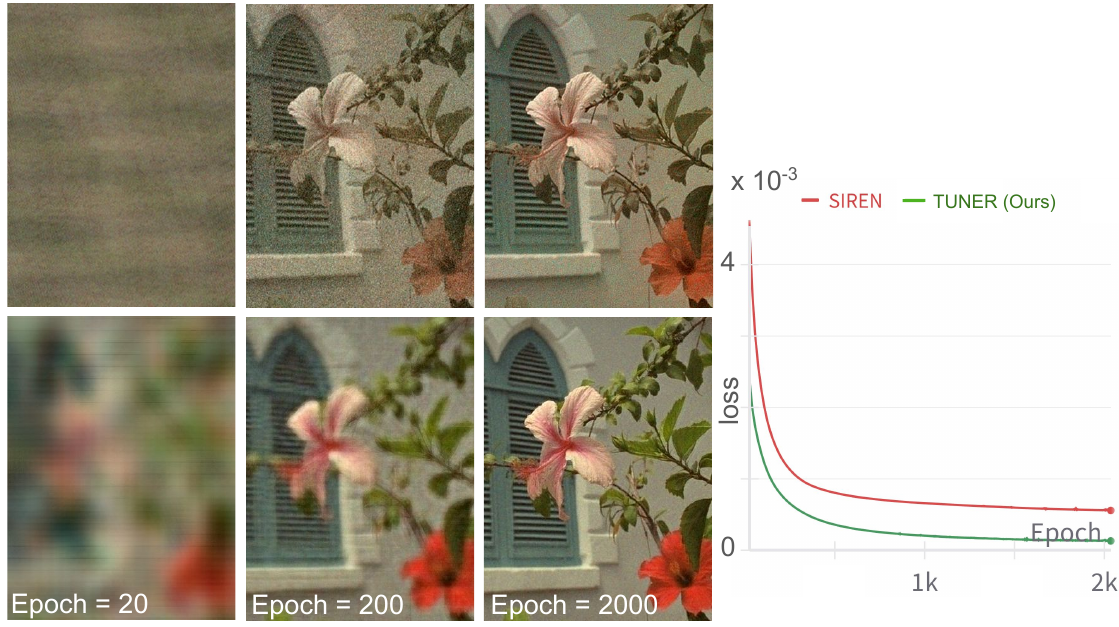}
    \vspace{-0.6cm}
    \caption{Comparing SIREN (top) and TUNER (bottom) 
    when we only know the Nyquist limit {(i.e, $\gt{b}\!=\!B$). Since $\gt{b}$ is big,} 
    % . Note that 
    SIREN introduces high frequencies at the beginning of training, resulting in very noisy reconstructions. Conversely, ours starts training low frequencies, 
    % obtaining a faster convergence. 
    {converging faster and being robust to $\gt{b}$}.}
    \label{fig: spectrum_bound}
    \vspace{-0.2cm}
\end{figure}

We first compare our method (TUNER) with SIREN, which requires choosing a 
% frequency 
{threshold} parameter $\gt{b}$,\footnote{Denoted as $\omega_0$ in \cite{sitzmann2020implicit}.} a critical task since higher $\gt{b}$ implies noisy reconstructions. TUNER also uses $\gt{b}$ to control the frequency generation, however, it provides more stable training with better reconstruction. 
To demonstrate this, we perform a comparison by varying $\gt{b}$ and the number of input ($m$) and hidden ($n$) neurons.
First, consider the case where we have no information about the signal bandlimit besides the Nyquist limit~$B$. 
Fig~\ref{fig: spectrum_bound} presents a comparison between SIREN (top) and TUNER (bottom) with $\gt{b}=B$. Note that SIREN produces noisy reconstructions, while ours learns faster and does not suffer from noise. This is due to our bounding scheme, which limits the generation of high frequencies and provides stable training (Sec \ref{chap: Bound}). In contrast, SIREN initializes its parameters such that the spectrum is initially bounded but does not maintain this guarantee during training.

\begin{figure}[!h]
% \vspace{-0.3cm}
\centering
         \includegraphics[width=\columnwidth]{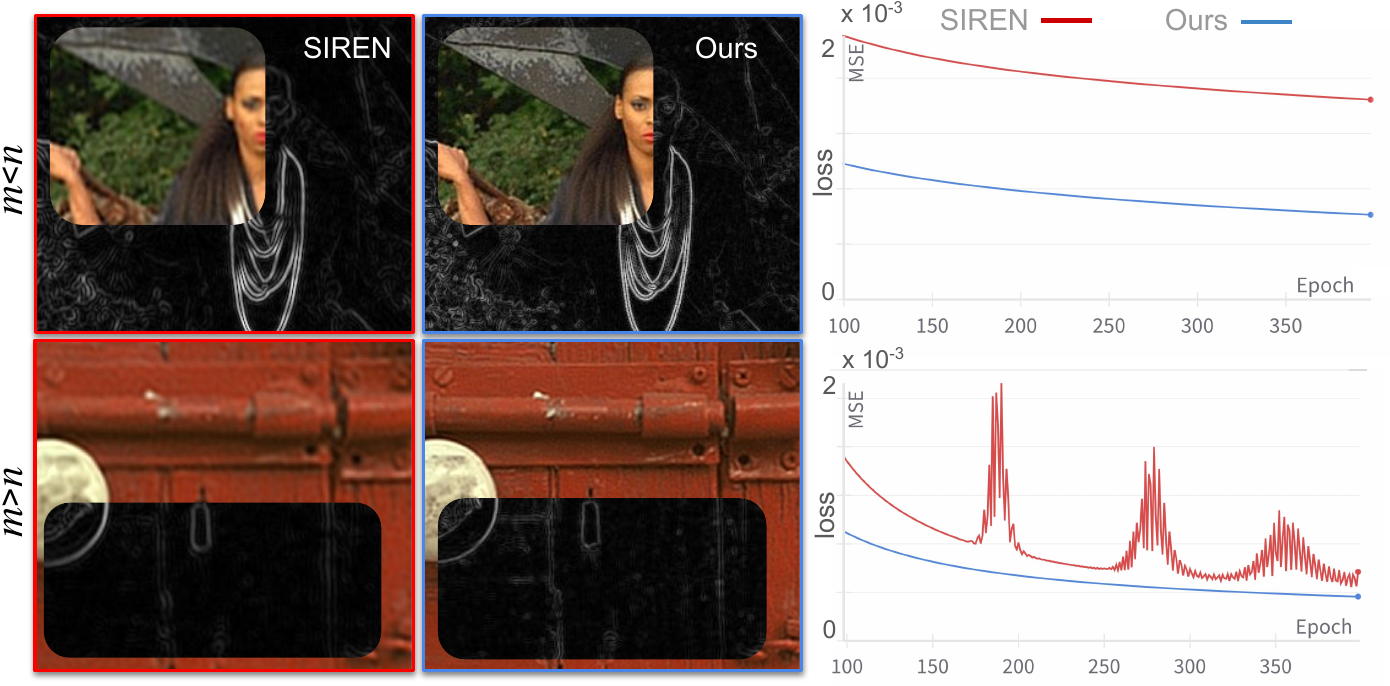}
        %  \vspace{-0.6cm}
    \caption{Comparison with SIREN when $\gt{b}=40$. TUNER improves training convergence in both cases ($m < n$, $m > n$) and  stability when $m > n$. {INR gradient shown in grayscale.}
    }
    \label{fig: siren_comparison}
    % \vspace{-0.2cm}
\end{figure}

Next, we consider a tuned $\gt{b}$ for SIREN to show that even under this condition, our method provides a better training scheme for sinusoidal INRs. Fig~\ref{fig: siren_comparison} presents the comparison varying the number of input ($m$) and hidden ($n$) neurons. We consider the cases $m<n$ (top) and $m>n$ (bottom), and TUNER performed better both in the signal/gradient reconstructions. Also, we note more robust training using TUNER, as shown in Fig~\ref{fig: siren_comparison}(right).

The previous experiment shows that initializing SIREN with a high $\gt{b}$ results in overfitting, while TUNER provides better reconstruction. On the other hand, with the same architecture and a tuned $\gt{b}$, our method also effectively learned the lower frequencies first and performed better (Fig~\ref{fig: siren_comparison}). 
% \andre{the previous sentences in this paragraph can be removed to save space, it was already clear before, no need to repeat}

Next, we numerically evaluate these experiments, providing a quantitative comparison with SIREN and BACON.
We compare different architectures ($m$, $n$) and spectrum bandlimits $\gt{b}$, evaluating the results in terms of signal and gradient PSNR over the supervised (Train) and unsupervised (Test) pixels.
Table \ref{tab: comparisons} presents these evaluations. 

\begin{table}[ht!]
% \small
\footnotesize
% \small{
\centering
\setlength{\tabcolsep}{1pt}
\begin{tabular}{cc|cccccc||cccccc}
\cline{3-14}
 &
   &
  \multicolumn{6}{c||}{Train} &
  \multicolumn{6}{c}{Test} \\ \cline{3-14} 
 &
   &
  \multicolumn{3}{c|}{Signal} &
  \multicolumn{3}{c||}{Gradient} &
  \multicolumn{3}{c|}{Signal} &
  \multicolumn{3}{c}{Gradient} \\ \hline
\multicolumn{1}{c|}{$\gt{b}$} &
  $(m,n)$ &
  \multicolumn{1}{c|}{\tiny SIREN} &
  \multicolumn{1}{c|}{\tiny BAC} &
  \multicolumn{1}{c|}{ours} &
  \multicolumn{1}{c|}{\tiny SIREN} &
  \multicolumn{1}{c|}{\tiny BAC} &
  ours &
  \multicolumn{1}{c|}{\tiny SIREN} &
  \multicolumn{1}{c|}{\tiny BAC} &
  \multicolumn{1}{c|}{ours} &
  \multicolumn{1}{c|}{\tiny SIREN} &
  \multicolumn{1}{c|}{\tiny BAC} &
  ours \\ \hline \hline
\multicolumn{1}{c|}{\multirow{3}{*}{85}} &
  1028,162 &
  \multicolumn{1}{c|}{\textbf{32.9}} &
  \multicolumn{1}{c|}{-} &
  \multicolumn{1}{c|}{\textbf{32.9}} &
  \multicolumn{1}{c|}{26.5} &
  \multicolumn{1}{c|}{-} &
  \textbf{28.8} &
  \multicolumn{1}{c|}{31.9} &
  \multicolumn{1}{c|}{-} &
  \multicolumn{1}{c|}{\textbf{32.2}} &
  \multicolumn{1}{c|}{25.5} &
  \multicolumn{1}{c|}{-} &
  \textbf{28.2} \\ %\cline{2-14} 
\multicolumn{1}{c|}{} &
  416,416 &
  \multicolumn{1}{c|}{34.0} &
  \multicolumn{1}{c|}{26.4} &
  \multicolumn{1}{c|}{\textbf{34.3}} &
  \multicolumn{1}{c|}{27.7} &
  \multicolumn{1}{c|}{22.4} &
  \textbf{28.9} &
  \multicolumn{1}{c|}{32.9} &
  \multicolumn{1}{c|}{25.4} &
  \multicolumn{1}{c|}{\textbf{33.4}} &
  \multicolumn{1}{c|}{27.1} &
  \multicolumn{1}{c|}{22.3} &
  \textbf{28.4} \\ %\cline{2-14} 
\multicolumn{1}{c|}{} &
  80,2024 &
  \multicolumn{1}{c|}{32.2} &
  \multicolumn{1}{c|}{-} &
  \multicolumn{1}{c|}{\textbf{33.1}} &
  \multicolumn{1}{c|}{26.6} &
  \multicolumn{1}{c|}{-} &
  \textbf{28.5} &
  \multicolumn{1}{c|}{31.3} &
  \multicolumn{1}{c|}{-} &
  \multicolumn{1}{c|}{\textbf{32.3}} &
  \multicolumn{1}{c|}{26.0} &
  \multicolumn{1}{c|}{-} &
  \textbf{28.0} \\ \hline
\multicolumn{1}{c|}{\multirow{3}{*}{171}} &
  1028,162 &
  \multicolumn{1}{c|}{31.3} &
  \multicolumn{1}{c|}{-} &
  \multicolumn{1}{c|}{\textbf{33.9}} &
  \multicolumn{1}{c|}{22.0} &
  \multicolumn{1}{c|}{-} &
  \textbf{27.2} &
  \multicolumn{1}{c|}{22.9} &
  \multicolumn{1}{c|}{-} &
  \multicolumn{1}{c|}{\textbf{32.9}} &
  \multicolumn{1}{c|}{20.7} &
  \multicolumn{1}{c|}{-} &
  \textbf{26.8} \\ %\cline{2-14} 
\multicolumn{1}{c|}{} &
  416,416 &
  \multicolumn{1}{c|}{29.2} &
  \multicolumn{1}{c|}{29.3} &
  \multicolumn{1}{c|}{\textbf{34.4}} &
  \multicolumn{1}{c|}{17.4} &
  \multicolumn{1}{c|}{26.7} &
  \textbf{27.0} &
  \multicolumn{1}{c|}{26.7} &
  \multicolumn{1}{c|}{26.2} &
  \multicolumn{1}{c|}{\textbf{33.1}} &
  \multicolumn{1}{c|}{16.2} &
  \multicolumn{1}{c|}{26.3} &
  \textbf{26.4} \\ %\cline{2-14} 
\multicolumn{1}{c|}{} &
  80,2024 &
  \multicolumn{1}{c|}{26.6} &
  \multicolumn{1}{c|}{-} &
  \multicolumn{1}{c|}{\textbf{30.8}} &
  \multicolumn{1}{c|}{14.2} &
  \multicolumn{1}{c|}{-} &
  \textbf{22.6} &
  \multicolumn{1}{c|}{23.8} &
  \multicolumn{1}{c|}{-} &
  \multicolumn{1}{c|}{\textbf{29.5}} &
  \multicolumn{1}{c|}{13.8} &
  \multicolumn{1}{c|}{-} &
  \textbf{21.3} \\ \hline
\multicolumn{1}{c|}{\multirow{3}{*}{256}} &
  1028,162 &
  \multicolumn{1}{c|}{26.8} &
  \multicolumn{1}{c|}{-} &
  \multicolumn{1}{c|}{\textbf{33.3}} &
  \multicolumn{1}{c|}{12.8} &
  \multicolumn{1}{c|}{-} &
  \textbf{24.7} &
  \multicolumn{1}{c|}{24.1} &
  \multicolumn{1}{c|}{-} &
  \multicolumn{1}{c|}{\textbf{31.9}} &
  \multicolumn{1}{c|}{12.7} &
  \multicolumn{1}{c|}{-} &
  \textbf{23.9} \\ %\cline{2-14} 
\multicolumn{1}{c|}{} &
  416,416 &
  \multicolumn{1}{c|}{24.7} &
  \multicolumn{1}{c|}{27.7} &
  \multicolumn{1}{c|}{\textbf{41.2}} &
  \multicolumn{1}{c|}{12.9} &
  \multicolumn{1}{c|}{\textbf{24.2}} &
  23.0 &
  \multicolumn{1}{c|}{22.3} &
  \multicolumn{1}{c|}{22.7} &
  \multicolumn{1}{c|}{\textbf{33.2}} &
  \multicolumn{1}{c|}{13.2} &
  \multicolumn{1}{c|}{\textbf{22.8}} &
  21.2 \\ %\cline{2-14} 
\multicolumn{1}{c|}{} &
  80,2024 &
  \multicolumn{1}{c|}{25.1} &
  \multicolumn{1}{c|}{-} &
  \multicolumn{1}{c|}{\textbf{30.0}} &
  \multicolumn{1}{c|}{13.0} &
  \multicolumn{1}{c|}{-} &
  \textbf{15.2} &
  \multicolumn{1}{c|}{24.1} &
  \multicolumn{1}{c|}{-} &
  \multicolumn{1}{c|}{\textbf{27.1}} &
  \multicolumn{1}{c|}{13.1} &
  \multicolumn{1}{c|}{-} &
  \textbf{14.6} \\ \hline
\end{tabular}
% \vspace{-0.2cm}
\caption{Comparison between SIREN, BACON (BAC), and our method. We use several architectures ($m$, $n$), bandlimits ($\gt{b}$), and evaluate over the train (90\%) and test(10\%) sets of signal/gradient.}
\label{tab: comparisons}
% \vspace{-0.4cm}
\end{table}

As expected, SIREN performs well with $\gt{b}=\frac{B}{3}$. In this case, TUNER is perceptually similar to SIREN but improves gradient reconstruction and consistently outperforms BACON.
When $\gt{b}=171$, TUNER offers superior quality in both signal and gradient, showing greater robustness to noise and superior performance even with $\gt{b}\geq\nicefrac{B}{3}$. Finally, with $\gt{b}=256$, our method surpasses SIREN in all cases. For BACON, our method achieves better signal reconstruction, with comparable gradients. However, the following experiment shows that BACON's reconstruction has visible defects in this scenario.
Note that we do not supervise the training with gradient information.

% {orange}{Experiments:

% % 1. SIREN VS Ours (the best config) VS BACON

% % 2. Compare BACON and Ours (hard vs soft) - bounding - bacon truncate the spectrum, ours do not.

% % 3. Improve Figure 10, using spectrum also, and maintain parameters in an appropriated range.

% Vary resolutions.
% }
% {orange}{Comapre bacon with 0 hidden layers}

\begin{figure}[h!]
\centering
    \includegraphics[width=0.95\columnwidth]{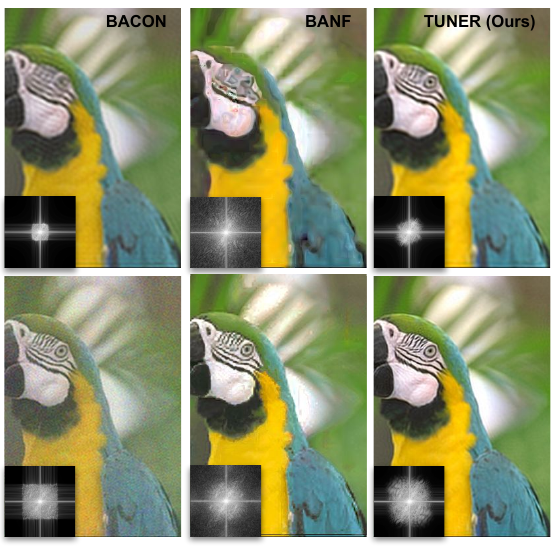}
    \vspace{-0.3cm}
    \caption{Comparison between BACON, BANF, and TUNER, trained with bandlimits $\gt{b}=60, 130$. Note that BACON uses a box filter, generating ringing artifacts (left). BANF (middle) has artifacts near edges and ambiguous bandlimit. Then, TUNER (right) resembles a soft filter, improving quality.}
    \label{fig: bacon_ours}
    \vspace{-0.4cm}
\end{figure}

\paragraph{Bandlimit comparison:}
Lastly, Fig~\ref{fig: bacon_ours} presents a qualitative comparison between BACON, BANF, and TUNER (Ours) with bandlimits $\gt{b}=60, 130$.
We train all methods with $~67$K parameters. For BANF we use the linear filter and adjust grid resolution to match the bandlimits and the MLP size to approximate $67$K parameters.
Particularly, the spectra were computed with the same process for all three methods.
Note that BACON truncates the spectrum with a hard filter, resulting in ringing artifacts (low bandlimit) and noisy reconstruction (high bandlimit).
BANF shows distortions around the eye area for both filters, and the spectra show no clear bandlimit. 
Conversely, TUNER operates as a soft filter providing better reconstructions for all cases.

\vspace{-0.4cm}
% \tnote{green}{improve explanation.}

\section{Conclusions and limitations}
\label{chap: conclusion}
% \tnote{green}{update conclusions}

We presented a study based on Fourier theory for sinusoidal MLPs, which resulted in novel, robust schemes for both initialization and spectral control of such networks. 
While our experiments focused on images to validate our theoretical claims, future work includes applying our schemes to other signal types. 
Additionally, we have focused our studies on 3-layer MLPs, leaving the investigation of deep networks for future research. We consider this work a first step toward unraveling the powerful expressiveness of sinusoidal~MLPs.

\paragraph*{Acknowledgments}
We would like to thank CAPES, FAPERJ, and Google
for partially funding this work.

{
    \small
    \bibliographystyle{ieeenat_fullname}
    \bibliography{refs}
}


\end{document}

\maketitle

\tableofcontents

\section{Theoretical analysis}

This section presents the proofs of Theorems 1 and 2 enunciated in the main paper, and a generalization of Theorem 1 for deep sinusoidal MLPs.
Then, we use the sine-cosine expansion of a sinusoidal INR (Equation (4) in the main paper) to obtain a closed formula for the coefficients of its Fourier series.
Finally, we analyze some representation problems (see Sec~4.1 in the main paper) that may arise due to arbitrary initialization, and present a simple but effective solution to avoid them.

\subsection{Proofs of Theorems 1 and 2}
\label{a-proofs}
% \tnote{red}{use the same notation..}
To prove Theorems 1 and 2 we consider the INR input to be 1D; the general case is analogous. 
Let $f({x})= \textbf{C}\circ\textbf{S}\circ\textbf{D}({x})+e$ be a sinusoidal INR. 
Here, the {input layer} $\textbf{D}({x})\!=\!\sin(\omega{x}+\varphi)$ projects the input ${x}$ into a list of harmonics with {frequencies} $\omega=(\omega_1,\ldots, \omega_{m})\in \R^{m}
$ and {shifts} $\varphi\in\R^{m}$.
Then, the sinusoidal layer $\textbf{S}(\cdot) = \sin(\textbf{W}\cdot + \textbf{b})$ with hidden matrix $\textbf{W} \in \R^{n\times m}$ and bias $\textbf{b} \in \R^{n}$ modulates those harmonics.
Finally, those neurons are combined using $\textbf{C}\cdot+e$, an affine transformation.
% Without loss of generality we assume $f$ to be 1D signal.
The $i$th hidden neuron can be written as follows: 
\begin{align}
h_i({x})
=\sin\left(
    \sum_{j=1}^m W_{ij}\sin(
        \omega_j{x}+\varphi_j
    )+ b_i
\right),    
\end{align}
where $W_{ij}$ are the $ij$ coefficients of $\textbf{W}$.
Note that $h_i$ receives a list of $m$ {input neurons} $\sin(\omega_j{x}+\varphi_j)$ that are combined with the weights $W_{ij}$ and activated by $\sin$. 
Before presenting the expansion of $h_i$, let us recall the Fourier series of a neuron with width $1$ and no bias~\cite[Page 361]{abramowitz1964handbook}:
\begin{align}\label{e-fourier-expansion-composition}
    \sin\big(W_{11}\sin(x)\big)\!=\!\!\sum_{k\in\Z \text{ odd}} \!\!\!J_k(W_{11}) \sin(kx),\,\,\text{with}\,\, J_k(W_{11})\!=\!\!\frac{1}{\pi}\int_0^\pi\!\!\!\cos\big(kt-W_{11}\sin(t)\big) dt.
\end{align}
The functions $J_k$ are the \textit{Bessel functions} of the first kind.
Theorem~\ref{thm: thm-1} provides an expansion for $h_i$ which generalizes \eqref{e-fourier-expansion-composition}. 
\begin{theorem}\label{thm: thm-1}
    Each hidden neuron $h_i$ of a $3$-layer sinusoidal MLP has an amplitude-phase expansion of the form
    \emph{
    \begin{align}\label{eq: thm-1}
        h_i({x}) = 
            \sum_{\textbf{k}\in\Z^{m}} \alpha_{\textbf{k}}\,\sin\big(\beta_{\textbf{k}}\,{{x}} + \lambda_{\textbf{k}}\big),
    \end{align}}
    \noindent where \emph{$\beta_{\textbf{k}} \!=\! \dot{\textbf{k}}{\omega}$}, \emph{$\lambda_{\textbf{k}}\!=\!\dot{\textbf{k}}{\varphi} \!+\! b_i$}, and  \emph{$\alpha_{\textbf{k}} \!=\! \prod_j\!J_{k_j}(W_{ij})$} is the product of the Bessel functions of the first kind.
\end{theorem}

\noindent To prove Theorem~\ref{thm: thm-1} we use the following lemma which will also be helpful for the generalization presented in next section.
\begin{lemma}
\label{a-lemma}
    Given \emph{$\textbf{a}=(a_1,\ldots, a_m), \textbf{y}=(y_1,\ldots, y_m)\in\R^m$} and \emph{$b\in\R$}, we have that
    \emph{
    \begin{align}\label{e-expansion-y}
        \sin\left( \sum_{i=1}^m{a}_i\sin({y}_i) + b \right) = \sum_{\textbf{k}\in\Z^m}\alpha_\textbf{k}\,\sin\Big( \dotprod{\textbf{k}, \textbf{y}} + b \Big),\quad \text{with} \quad \alpha_\textbf{k}=\prod_{l=1}^mJ_{k_l}(a_l).
    \end{align}}
\end{lemma}

\begin{proof}
The proof consists of verifying \eqref{e-expansion-y} as well as a similar formula using the cosine as the activation function:
\begin{align}\label{e-perceptron_approx_cosine}
\cos\left(
    \sum_{i=1}^m a_{i}\sin(
y_i
    )+ b
\right)
= \sum_{\textbf{k}\in\Z^m}\alpha_\textbf{k}\cos\Big(
    \dot{\textbf{k}}{\textbf{y}}+ b
\Big).
\end{align}

The proof is by induction in $m$. For the \textbf{}base case $m=1$, we prove $\sin\left(a_1 \sin(y_1)+b\right)=\sum_{k\in \Z} J_k(a_1) \sin(ky_1+b)$.
For this, we use \eqref{e-fourier-expansion-composition} and its cosine analogous expansion $\cos\big(a_1\sin(y_1)\big)\!=\!\!\sum J_l(a_1) \cos(ly_1)$, here the sum is over the even numbers. Thus, by the angle sum identity we obtain:
\begin{align*}
    \sin\big(a_1 \sin(y_1)+b\big)&=\sin\big(a_1 \sin(y_1)\big)\cos(b)+\cos\big(a_1 \sin(y_1)\big)\sin(b)\\
    &=\sum_{k\in\Z \text{ odd}} \!\!\!J_k(a_1) \sin(ky_1)\cos(b) + \sum_{l\in\Z \text{ even}} \!\!\!J_l(a_1) \cos(ly_1)\sin(b)\\
    &=\sum_{k\in\Z \text{ odd}} \!\!\!J_k(a_1) \sin(ky_1+b) + \sum_{l\in\Z \text{ even}} \!\!\!J_l(a_1) \sin(ly_1+b)\\
    &=\sum_{k\in\Z} J_k(a_1) \sin(ky_1+b).
\end{align*}
In the third equality we combined the formula $\sin(u)\cos(v)=\frac{\sin(u+v)+\sin(u-v)}{2}$ and the fact that $J_{-k}(u)=(-1)^kJ_k(u)$ to rewrite the summations.
The proof of the formula using the cosine as an activation function is similar.

Assume that the formulas hold for $m-1$, with $m>1$, we prove that \eqref{e-expansion-y} holds for $m$ (the \textbf{induction step}). 
% Again, we denote $y_j=\omega_j {x}+ \varphi_j$ and $b_i$ as $b$ for simplicity.

\begin{align}
    \sin\left(\sum_{j=1}^{m} a_j\sin(y_j)+ b\right)&=\sin\left(\sum_{j=1}^{m-1} a_j\sin(y_j)+ b\right)\cos\big(a_m\sin(y_m)\big)\\
    &+\cos\left(\sum_{j=1}^{m-1} a_j\sin(y_j)+ b\right)\sin\big(a_m\sin(y_m)\big)
    \\
    &=\sum_{\textbf{l}\in\Z^{m-1},\, k\in\Z \text{ even}}\alpha_\textbf{l}J_k(a_m)\sin\Big(\dot{\textbf{l}}{\textbf{y}}+ b\Big)\cos(ky_m)\\
    &+\sum_{\textbf{l}\in\Z^{m-1},\, k\in\Z \text{ odd}}\alpha_\textbf{l}J_k(a_m)\cos\Big(\dot{\textbf{l}}{\textbf{y}}+ b\Big)\sin(ky_m)\\
    &=\sum_{\textbf{k}\in\Z^m}\alpha_\textbf{k} \sin\Big(\dot{\textbf{k}}{\textbf{y}}+ b\Big)\label{e-lema}
\end{align}
We use the induction hypothesis in the second equality and an argument similar to the one used in the base case to rewrite the harmonic sum. Again, the cosine activation function case is analogous.
\end{proof}

Note that the proof of Theorem \ref{thm: thm-1} is a particular case of Lemma~\ref{a-lemma} with ${\textbf{a}}=\textbf{W}_{i}$, ${\textbf{y}} = \omega \textbf{x} + \varphi$, and $b = b_i$; where $\textbf{W}_{i}$ is the $i$th row of $\textbf{W}$.
\citet{yuce2022structured} presented a similar formula for MLPs activated by polynomial functions.
While it is evident that the sine function can be approximated by a polynomial using Taylor series, our formula requires no approximations.
In addition to providing a simple proof, we also derive the analytical expressions for the amplitudes. These expressions enable us to compute upper bounds (Theorem~2) for the new frequencies $\alpha_\textbf{k}$ in terms of $\textbf{k}$ and $\W$.
% Next, we prove such inequality,
% \begin{align*}
%     \norm{\alpha_\textbf{k}}=\prod_{j=1}^m\Big( \tfrac{|W_{ij}|}{2}\Big)^{|k_j|}\frac{1}{|k_j|!}.
% \end{align*}

\begin{theorem}\label{thm: bound}  %CORRECT?
    The magnitude of the amplitudes \emph{$\alpha_\textbf{k}$} in the expansion \eqref{eq: thm-1} is bounded by
    \emph{$
        \prod_{j=1}^m\Big( \tfrac{|W_{ij}|}{2}\Big)^{|k_j|}\frac{1}{|k_j|!}.
    $}
    % \vspace{-0.2cm}
\end{theorem}
\begin{proof}
Theorem~1 says that
$\alpha_\textbf{k}=\prod_{j=1}^{m}J_{k_j}(W_{ij}).$
To estimate an upper bound for this number, we use the following inequality~\cite{paris1984inequality}, which gives an upper bound for the Bessel functions $J_k$.
\begin{align}\label{e-upper_bound_bessel_functions}
    \abs{J_k(W_{ij})}<\frac{\left(\frac{\abs{W_{ij}}}{2}\right)^k}{k!}\,, \quad k>0,\quad W_{ij}>0.
\end{align}
Observe that this inequality also holds for $W_{ij}<0$ since $\abs{J_k(-u)}=\abs{J_k(u)}$.
Therefore, replacing \eqref{e-upper_bound_bessel_functions} in $\prod_{j=1}^{m}J_{k_j}(W_{ij})$ and using $\abs{J_{-k}(W_{ij})}=\abs{J_k(W_{ij})}$ results in the desired inequality.
\end{proof}

\subsection{Extension to deeper networks}
\label{a-extension}

We extend Theorem~\ref{thm: thm-1} to deeper neurons using a recursive argument. 
For this, let $f$ be a sinusoidal MLP with $d>1$ hidden layers defined as $f(\textbf{x}) = \textbf{C}\circ\textbf{S}_{d}\circ\cdots\circ\textbf{S}_1\circ\textbf{S}_0(\textbf{x})+e$ where $\textbf{S}_i(\textbf{x}) = \sin\big(\textbf{W}^{i}\textbf{x} + \textbf{b}^{i}\big)$ is a sinusoidal layer with hidden weights $\W^{i}\in\R^{n_{i+1}\times n_{i}}$ and bias $\textbf{b}^{i}\in\R^{n_{i+1}}$. 
For simplicity, we denote $\W^0:=\omega$ and $\textbf{b}^0:=\varphi$.

Now, the $j$th neuron of the $i$th hidden layer $h_j
^i$ can be written as 
\begin{align}\label{eq: neuron}
h_j^{i}(\textbf{x})=\sin\left(\sum_{k}{\W}_{jk}^{i}\underbrace{\sin(\textbf{y}^{i-1})}_{\textbf{h}^{i-1}(\textbf{x})}+b_j^{i}\right),    
\end{align}
where $\textbf{y}^{i-1}$ is the linear part of $\textbf{h}^{i-1}(\textbf{x})$. 
Again, we prove that \eqref{eq: neuron} has an expansion which generalizes \eqref{eq: thm-1} to deeper networks.

\begin{theorem}\label{t-generalization}
    Each hidden neuron \emph{$h^i_j(\textbf{x})$} of a sinusoidal INR has the following expansion,
    \emph{\begin{align}\label{eq: generalization}
        h_j^i(\textbf{x}) = 
            \sum_{\textbf{k}\in\,\Z^{n_{1}+\cdots+ n_i}}\! \!\!\!\!\!\!\!\!\alpha\,\sin\big(\dotprod{\textbf{k}^1, \omega}\textbf{x} + \lambda \big),
    \end{align}}
    with \emph{$\textbf{k}=\big(\textbf{k}^1,\cdots,\textbf{k}^i\big)$} being a tuple of integer vectors, \emph{$\displaystyle\lambda = {b}^{i}_j+ \sum_{l=1}^{i}\dotprod{\textbf{k}^{l}, \textbf{b}^{l-1}}$}, and 
    \emph{$\displaystyle\alpha = 
    \alpha_{i}(\textbf{W}^{i}_{j})
    \prod_{l=1}^{i-1}
    \alpha_{l}(\textbf{k}^{l+1}\textbf{W}^{l})
    $}
    where \emph{$\alpha_{l}(\textbf{a}) := J_{k^l_1}(a_1)\cdots J_{k^l_{n_l}}(a_{n_l})$} is the product of the Bessel functions of the first kind and $\W^i_j$ is the $j$th row of $\W^i$.
\end{theorem}
\begin{proof}
The proof is by induction on the depth $i$ of the neuron $h^i_j(\x)$.
First, note that the \textbf{base case} ($i=1$) is Theorem~\ref{thm: thm-1}.
For the \textbf{induction step}, we suppose the hypothesis holds for depth $i-1>0$ and prove that it holds for $i$. In this case, we have that the $j$th neuron of the $i$th layer is given by,
$$h_j^i(\textbf{x}) = \sin\Big(\W_{j}^i \sin\big(\W^{i-1}\textbf{h}^{i-2}(\x)+\textbf{b}^{i-1}\big)+b^i_j\Big).$$ Considering $\textbf{a} :=\W_{j}^i$,\, $\textbf{y}:=\W^{i-1}\textbf{h}^{i-2}(\x)+\textbf{b}^{i-1}$ and $b:=b^i_j$, Lemma~\ref{a-lemma} implies that
\begin{align*}
    h^i_j(\x) & = \!\!\!\sum_{\textbf{k}^{i}\in\Z^{n_{i}}}\!\!\!\! \alpha_{i}(\textbf{a}) \,\sin\left(\dotprod{\textbf{k}^{i}, \textbf{y}} + b\right) \\
    & = \!\!\!\sum_{\textbf{k}^i\in\Z^{n_{i}}}\!\!\!\!\alpha_i \left(\W_{j}^{i}\right)\underbrace{\sin\Big(\dotprod{\textbf{k}^i, \W^{i-1}\textbf{h}^{i-2}\!(\x)\!+\!\textbf{b}^{i-1}}\!+\! b_j^{i}\Big)}_{\widetilde{{h}}^{i-1}(\x)}\tag{\textasteriskcentered},
    % & = \!\!\!\sum_{\textbf{k}^i\in\Z^{n_{i}}}\!\!\!\!\!\!\alpha_i\!\!\left(\W_{j*}^{i}\right) \sin\!\!\Big( \textbf{k}^i\W^{i-1}\textbf{h}^{i-2}\!(\x)\!+\!\dotprod{\textbf{k}^i,\textbf{b}^{i-1}}\!+\! b_j^{i}\Big).
\end{align*}
where $\widetilde{{h}}^{i-1}(\x)$ is a list of hidden neurons with weights $\widetilde{\W}:=\textbf{k}^i\W^{i-1}$ and bias $\widetilde{b}:=b^i_j + \dotprod{\textbf{k}^i, \textbf{b}^{i-1}}$:
$$
\widetilde{{h}}^{i-1}(\x) = \sin\Big(\widetilde{\W} 
\sin\big(\W^{i-2}\textbf{h}^{i-3}(\x)+\textbf{b}^{i-2}\big) +\widetilde{b}\Big).
\footnote{For $i=2$, we define $\textbf{h}^{i-3}(\x)=\x$.}
$$
Since $\widetilde{{h}}^{i-1}$ has depth $i-1$, the induction hypothesis implies 
\begin{align}\label{induction-step}
    \widetilde{{h}}^{i-1}(\x) = \sum_{\textbf{k}\in\Z^{n_1+\cdots+ n_{i-1}}}\alpha\sin\Big(\dotprod{\textbf{k}^1, \W^0}\x + \lambda \Big),
\end{align} 
where $\alpha$ and $\lambda$ are defined as follows: 
\begin{align*}
\alpha=\alpha_{i-1}(\widetilde{\W})\prod_{l=1}^{i-2}
    \alpha_{l}(\textbf{k}^{l+1}\textbf{W}^{l}) = \prod_{l=1}^{i-1}\alpha_{l}(\textbf{k}^{l+1}\textbf{W}^{l}),\quad
    \lambda = \widetilde{b}+\sum_{l=1}^{i-1}\dotprod{\textbf{k}^{l}, \textbf{b}^{l-1}} = b^{i}_j+\sum_{l=1}^{i}\dotprod{\textbf{k}^{l}, \textbf{b}^{l-1}}.
\end{align*}
Replacing \eqref{induction-step} in Equation (\textasteriskcentered), we obtain the desired expression:
\begin{align*}
h^{i}_j(\x) &= \sum_{\textbf{k}^i\in\Z^{n_{i}}}\alpha_i(\W_{j}^{i})\sum_{\textbf{k}\in\Z^{n_1+\cdots+ n_{i-1}}}\alpha \, \sin\left(\dotprod{\textbf{k}^0, \W^0}\x + \lambda \right)\\
    &= \sum_{\textbf{k}\in\Z^{n_1+\cdots+n_{i}}}\big(\alpha_i(\W_{j}^{i})\,\alpha\big)\,\sin\left(\dotprod{\textbf{k}^0, \W^0}\x + \lambda \right).
\end{align*}
\end{proof}

Note that several properties observed in INRs with a single hidden layer also extend to the general case. First, the amplitudes \(\alpha\) depend only on the hidden matrices and remain products of Bessel functions. Second, the biases determine the shifts \(\lambda\), suggesting that not all of them may be necessary during training.  
Furthermore, we emphasize that adding more hidden layers does not introduce new frequencies, as these are still given by \(\dotprod{\textbf{k}, \omega}\). Instead, deep sinusoidal networks primarily refine the amplitudes of the generated frequencies, which are entirely determined by the choice of \(\omega\).

\subsection{Towards computing the Fourier series of a sinusoidal MLP}
\label{a-fourier-transform}
Without loss of generality, we will assume the $3$-layer sinusoidal MLP $f$ (defined in Section~\ref{a-proofs}) has $\R^2$ as its domain. To compute its Fourier series, we recall that Equation (4) from the main paper states that $f$ can be rewritten as
\begin{align}
\label{eq: inr-expansion}
   f(\textbf{x}) =\sum_{\textbf{k}\in\Z^m}\dotprod{\textbf{C},{A}_{\textbf{k}}}\sin\big(\beta_{\textbf{k}}\textbf{x}\big) + \dotprod{\textbf{C}, {B}_{\textbf{k}}} \cos\big(\beta_{\textbf{k}}\textbf{x}\big) + e,
\end{align}
where $\beta_{\textbf{k}}=\dot{\textbf{k}}{\omega}=\textbf{k}^{\top}\omega$. This expression resembles a Fourier series, however, a given frequency $\mathcal{F}\in\Z^2$ may appear several times as generated frequencies $\beta_\textbf{k}$ associated with different coefficients $\textbf{k}$.
Here, we provide a characterization of all $\textbf{k}\in\Z^m$ such that $\textbf{k}^{\top}\omega\!=\!\frac{2\pi}{p}\mathcal{F}$, where $p$ is the INR period, leading to an algorithm to compute the Fourier series of $f$.

To guarantee that $f$, initialized with $\omega= \frac{2\pi}{p}\left[\textbf{f}^{\,x}, \textbf{f}^{\,y}\right]\in\frac{2\pi}{p}\Z^{m\times2}$, can represent an arbitrary signal, we must determine when it can reconstruct any given frequency $\mathcal{F}\in\Z^2$.
This is equivalent of finding a solution $\textbf{k}$ to the following system of Diophantine equations:
\begin{equation}\label{eq: diophantine}
\left[\textbf{f}^{\,x}, \textbf{f}^{\,y}\right]^\top\textbf{k} = \mathcal{F} \quad \text{for any} \quad \mathcal{F}\in\Z^2.
\end{equation}
To solve \eqref{eq: diophantine} we use the approach in \cite[p. 50]{zippel2012effective}. Specifically, we consider the Smith normal form of $\left[\textbf{f}^{\,x}, \textbf{f}^{\,y}\right]$, that is, $\textbf{B}= \textbf{U}\left[\textbf{f}^{\,x}, \textbf{f}^{\,y}\right]^\top\textbf{V}$, where $\textbf{U}\in\Z^{2\times2}$ and $\textbf{V}\in\Z^{m\times m}$ are inversible matrices. 
Defining $\textbf{e}=\textbf{U}\mathcal{F}$,  there exist solution for \eqref{eq: diophantine} only if $B_{ii}$ divides $e_i$ for all $i$.
Thus, to have integer solutions for \eqref{eq: diophantine} we need $e_i/B_{ii}\in\Z$. Finally, the solutions have the form
$\textbf{V}\left[\begin{smallmatrix}
\frac{e_1}{B_{11}} & \frac{e_2}{B_{22}} & l_{1} & ... & l_{m-2} \end{smallmatrix}\right]^T$
with $l_{1},\ldots, l_{m-2}$ arbitrary integers.

Now, recall that we are considering $\omega_1=\frac{2\pi}{p}(1,0)$ and $\omega_2=\frac{2\pi}{p}(0,1)$. Then, the Smith normal form and unimodular matrices $\textbf{U}$, \textbf{V} are reduced to, 
$$
\textbf{B} = \left[\id{2} \,|\, \textbf{0}_{2\times m-2}\right],\quad \textbf{U}=\mathbbm{I}_2, \quad\text{and}\quad
\textbf{V} = 
\begin{bmatrix}
 \id{2} & \vline &
\begin{matrix}    
-\text{f}^{\,x}_{3} & ... & -\text{f}^{\,x}_{m} \\
-\text{f}^{\,y}_{3} & ... & -\text{f}^{\,y}_{m} \\
\end{matrix} \\
\hline
\textbf{0}_{m-2 \times 2} & \vline & \id{m-2}
\end{bmatrix}.
$$
Since $B_{ii}=1$, the divisibility condition is satisfied. Additionally, we have $\textbf{e}\!=\!\mathcal{F}$ which implies that the set of integer solutions of \eqref{eq: diophantine} is given by
$$
\mathcal{C}_\mathcal{F} = 
\begin{Bmatrix}
    \begin{bmatrix} 
        \mathcal{F}_1 - l_1 \text{f}^{\,x}_{3} - \cdots - l_{m-2}  \text{f}^{\,x}_{m} \\
        % \sum_{k=1}^{m-2} l_k \text{f}^{\,x}_{k+2} \\
        \mathcal{F}_2 - l_1 \text{f}^{\,y}_{3} - \cdots - l_{m-2}  \text{f}^{\,y}_{m} \\
        % \sum_{k=1}^{m-2} l_k \text{f}^{\,y}_{k+2} \\
        l_1 \\
        \vdots \\
        l_{m-2}
    \end{bmatrix}: l_1, ..., l_{m-2} \in \Z
\end{Bmatrix}.
$$

We obtain the Fourier series of $f$ by aggregating all coefficients associated with each frequency $\mathcal{F}\in\Z^2$ in~\eqref{eq: inr-expansion}. That is, the Fourier coefficients are given by 
$\widehat{A}_{\mathcal{F}}=\sum_{\textbf{k}\in\mathcal{C}_\mathcal{F}}\dotprod{\textbf{C},{A}_{\textbf{k}}}$ and $\widehat{B}_{\mathcal{F}}=\sum_{\textbf{k}\in\mathcal{C}_\mathcal{F}}\dotprod{\textbf{C},{B}_{\textbf{k}}}$. 
% Note that for the frequency $(0,0)$, it's necessary to add the last bias $d$ to the coefficient $\widehat{B}_{(0,0)}$.
% To find a similar formula for $f$ with domain in $\R^d$ (i.e.,  $\omega\in\R^{m\times d}$), we just need to set $\omega_i=\textbf{e}_i$ and aggregate the coefficients analogously.

\subsection{On the sub-periodicity of sinusoidal MLPs}
\label{ap: subperiods}
Initializing a sinusoidal MLP $f$ using input neurons with period $p$ implies that $f$ is also periodic with period $p$, however, the initialization could generate sub-periods implying that we can not fit a signal with fundamental period~$p$.
Next, we derive a condition to avoid such a problem.
First, recall that Theorem~\ref{thm: thm-1} says that a neuron 
$h(\textbf{x}) =
\sin\big(
    \W_i \sin(
        \omega\x+\varphi
    ) +b_i
\big)$, with $\omega = \frac{2\pi}{p}\left[\textbf{f}^{\,x}, \textbf{f}^{\,y}\right]$ for some $\textbf{f}^{\,x}, \textbf{f}^{\,y}\in\Z^m$, can be expressed as:
\begin{align}\label{eq: expansion}
    h(\textbf{x}) = \sum_{\textbf{k}\in\Z^m}\alpha_\textbf{k} \sin\big(\beta_{\textbf{k}}\textbf{x} + \lambda_{\textbf{k}}\big)
\end{align}
with $\beta_{\textbf{k}}=\dotprod{\textbf{k}, \omega}$ and $\lambda_{\textbf{k}}= \dotprod{\textbf{k}, \varphi}+b_i$. Therefore,
$\beta_{\textbf{k}}\textbf{x} = \frac{2\pi}{p}\Big(\dotprod{\textbf{k}, \textbf f^{\,x}}x + \dotprod{\textbf{k}, \textbf f^{\,y}}y\Big).$
Now, suppose that $h$ has sub-period $\frac{p}{q}$ in $x$-axis and $\frac{p}{s}$ in $y$-axis ($q,s\in\Z^+$), i.e. $h(x,y)=h\left(x+\frac{p}{q}, y+\frac{p}{s}\right)$.
Then, \eqref{eq: expansion} implies
\begin{align}\label{eq: expansion2}h\bigg(x+\frac{p}{q}, y+\frac{p}{s}\bigg) = \sum_{\textbf{k}\in\Z^m}\alpha_\textbf{k} \sin\bigg(\beta_{\textbf{k}}\textbf{x} + \lambda_{\textbf{k}} + 2\pi\dotprod{\textbf{k}, \frac{\textbf{f}^{\,x}}{q}+\frac{\textbf{f}^{\,y}}{s}}\bigg).
\end{align}
Thus, since $\sin(x+2\pi k )=\sin(x)$ for $k\in\Z$, we have that~\eqref{eq: expansion} and \eqref{eq: expansion2} are equal only if $\dotprod{\textbf{k}, \frac{\textbf{f}^{\,x}}{q}+\frac{\textbf{f}^{\,y}}{s}}\in\Z$ for all $\textbf{k}\in\Z^m$. Therefore, $h$ (and consequently $f$) has sub-periods only if there exists $q>1$ or $s>1$ such that $\frac{\textbf{f}^{\,x}}{q}+\frac{\textbf{f}^{\,y}}{s}\in\Z^m$.
To avoid this problem, we define the first elements of $[\textbf{f}^{\,x}, \textbf{f}^{\,y}]$ as $\f^{\,x}_{1}=1$, $\f^{\,y}_{1}=0$, $\f^{\,x}_{2}=0$, and $\f^{\,y}_{2}=1$.
Note that this implies
$\frac{\f^{\,x}_{1}}{q}+\frac{\f^{\,y}_{1}}{s}=\frac{1}{q}\in\Z$ if and only if $q=1$, and $\frac{\f^{\,x}_{2}}{q}+\frac{\f^{\,y}_{2}}{s}=\frac{1}{s}\in\Z$ if and only if $s=1$.

\begin{figure}[!h]
     \centering
     \begin{subfigure}[t]{0.25\textwidth}
         \centering
         \includegraphics[width=\textwidth]{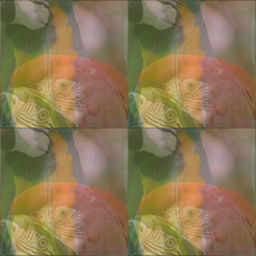}
         \caption{$\omega$ defined with even frequencies.}
     \end{subfigure}
     \begin{subfigure}[t]{0.25\textwidth}
         \includegraphics[width=\textwidth]{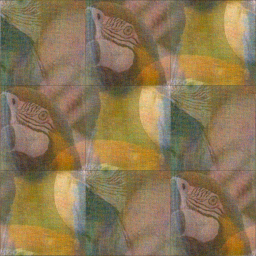}
         \caption{\footnotesize{$\omega\!\!=\!\!\{(u,v)|u,v\!\in \pm\big[\!1, \!10, \!100\big]\}$}}
     \end{subfigure}
     \begin{subfigure}[t]{0.25\textwidth}
         \centering
         \includegraphics[width=\textwidth]{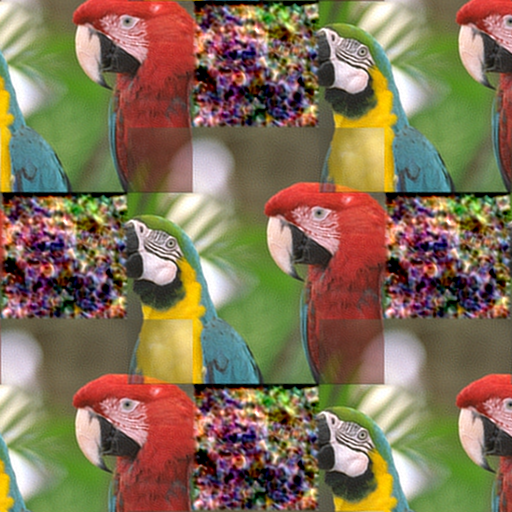}
         \caption{Extrapolation to $[-2, 2]^2$ of $f$ ($p\!=\!3$) with $\omega$ as in (a) .}
     \end{subfigure}
     \vspace{-0.3cm}
        \caption{Bad reconstructions given by specific (wrong) initializations of the input frequencies.}
        \label{fig: sub_periods}
        \vspace{-0.4cm}
\end{figure}

\vspace{0.2cm}
Fig~\ref{fig: sub_periods} shows cases where the initialization of $\omega$ does not hold the above condition resulting in poor reconstructions. We train networks with architecture $m=32$, $n=1500$ (number of hidden neurons), and bandlimit $\gt{b}=100$. We networks are trained for $3000$ epochs on a $256^2$ resolution image.
In Fig~\ref{fig: sub_periods}(a) and (b), we initialize $[\textbf{f}^{\,x}, \textbf{f}^{\,y}]$ using even frequencies and the cartesian product of $\{1, 10, 100\}$, with period $p=2$, respectively.
In Fig~\ref{fig: sub_periods}(c), we use the same initialization as in Fig\ref{fig: sub_periods}(a) but increase the period to $p=3$ and visualize an extrapolation of the training domain. This highlights the overlap of copies caused by the sub-periodicity due the poor initialization. 
We fix this by adding $(0,1),(1,0)$ to $\omega$, see Fig~3 in the main text.
% 
% \pagebreak
\section{Additional experiments}
% This section presents additional experiments to better understand our method. 
% First, present an ablation for the initialization of the input and hidden layers. 
% Then, we analyze the behavior of the learned bounds when initializing $\omega$ with several bandlimits $\gt{b}$.
% We also discuss the representational capacity of layer composition. 
% Finally, we extend the bandlimit comparison with BACON in Fig~1  and present an additional comparison with SIREN.
% 
% Again, our sinusoidal INR $f:\R^2\to\R^3$ is determined by $n, m$, the dimensions of its hidden matrix. 
% The bandlimit of the sampled signal is given by $B$, while $\gt{b}\leq B$ is the bandlimit for the input frequencies $\omega$.
% Specifically, we sample $\omega$ in $[-\gt{b}, \gt{b}]^2 = \textbf{L}\sqcup\textbf{H}$, where $\textbf{L}$ ($\textbf{H}$) is a region of low (high) frequencies.
% Our bounding scheme considers bounding the columns of $\W$ related to low (high) frequencies using the bounds $\gt{c}_{\textbf{L}}>0$ ($\gt{c}_{\textbf{H}}>0$).
% \subsection{Implementation details}
% Note that if $\omega_j$ is an input frequency, $-\omega_j$ is also in the spectrum. 

\subsection{Initialization}
\begin{figure}[!h]
\vspace{-0.2cm}
\centering
     \begin{subfigure}[b]{0.16\textwidth}
         \centering
         \includegraphics[width=\textwidth]{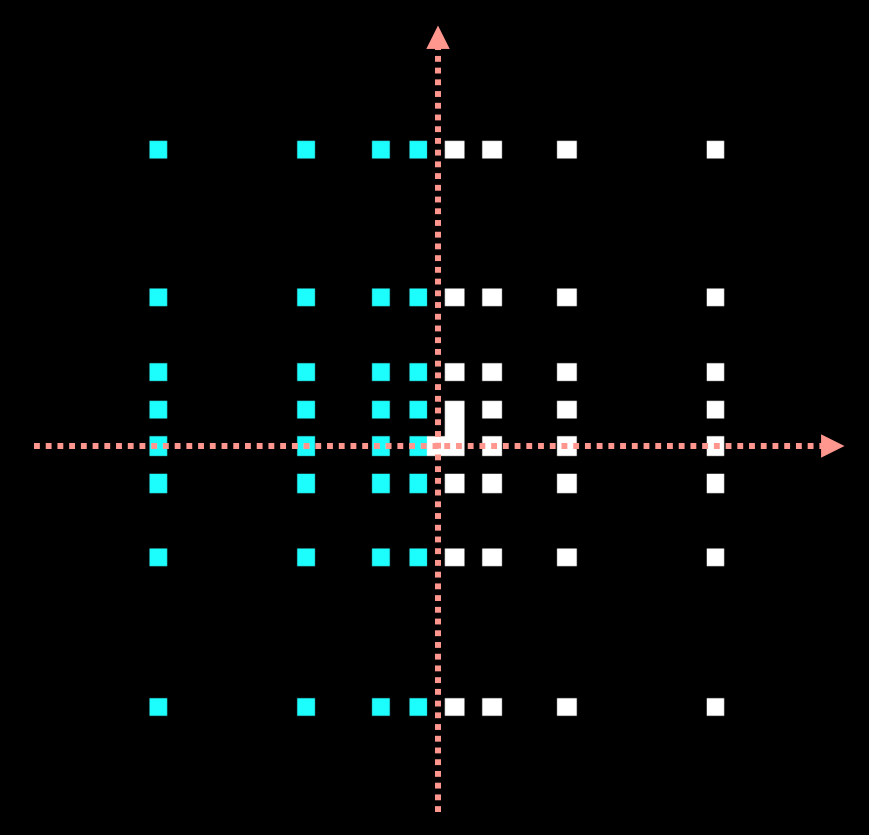}
         \caption{Initialization of $\omega$. Blue dots are the negative frequencies of those in white.}
     \end{subfigure}
     \quad
     \begin{subfigure}[b]{0.165\textwidth}
         \centering
         \includegraphics[width=\textwidth]{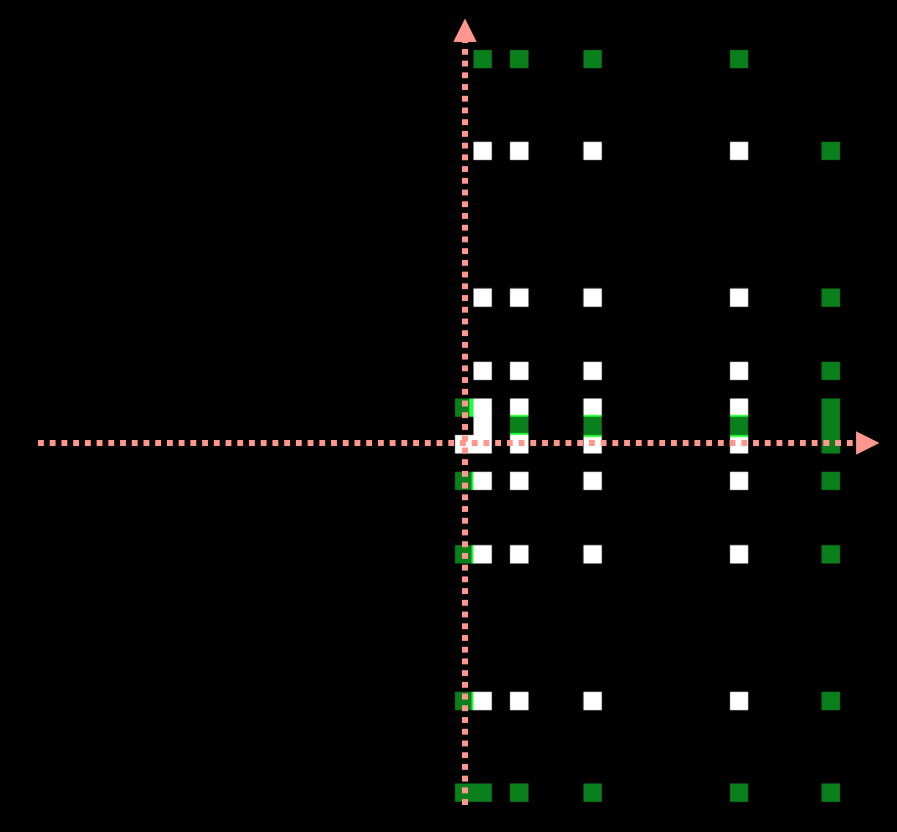}
         \caption{Initialization of $\omega'$. The frequencies in green replace those in blue.}
     \end{subfigure}
     \quad
     \begin{subfigure}[b]{0.29\textwidth}
         \centering
         \includegraphics[width=\textwidth]{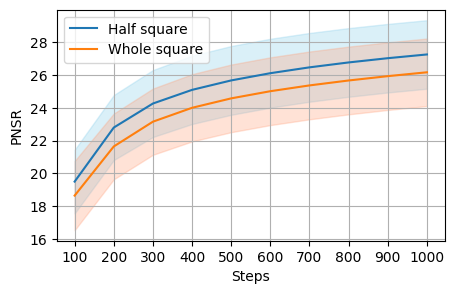}
         \caption{Comparison of PSNR during training.}
     \end{subfigure}
     \vspace{-0.2cm}
    \caption{Comparison of reconstructions with different input frequency initializations.
    }
    \label{fig: halfsquare}
\end{figure}

%   \begin{wrapfigure}[10]{r}{0.13\textwidth}
%   \centering\vspace{-.5cm}
%     \includegraphics[width=0.13\textwidth]{TUNER/suppmat_spectrum_partition.png}
%     % \caption{Partition of \textbf{K} in $\R^2$.}
%     % \label{fig: spectrum_partition}
% \end{wrapfigure}
Note that initializing a frequency $\omega_i$ implies that its negative $-\omega_i$ also appears in the spectrum.
This is a consequence of $\sin(\omega_j\x+\varphi_j)=\cos(\varphi_j)\sin(\omega_j\x)-\sin(\varphi_j)\sin\left(-\omega_j\x+\pi/2\right)$.
Then, we only need to sample in half of $[-\gt{b}, \gt{b}]^2$, with $\gt{b}$ being the bandlimit of the input frequencies, allowing the sampling of more frequencies.
We use this fact to avoid sampling duplicated frequencies to obtain a better reconstruction.  
We test the capacity of two INRs with the same architecture $m\!=\!102$, $n\!=\!512$, but with different initializations for $\omega$ (blue and white dots in Fig~\ref{fig: halfsquare}a) and $\omega'$ (white/green dots in Fig~\ref{fig: halfsquare}b):
\begin{align*}
    \omega =\big[(k,l)|\,\, k,l\in\{1, 3, 4, 7, 10, 20\}\big]\,\,\text{and}\,\,
    \omega' =\big[(k,l)|\,\, (k,l)\in\omega \text{ with } k>0\big] \sqcup \eta
\end{align*}

\noindent where $\eta$ are additional frequencies (in green) sampled in the Cartesian product of $[0, 1, 2, 3, 4, 7, 10, 20]$.

Fig~\ref{fig: halfsquare}(c) shows the PSNR of the resulting networks with respect to the iterations.
The results show that adding the new frequencies $\eta$ results in a slight but consistent improvement in the PSNR during training.
Also, from \eqref{eq: inr-expansion} the final bias $e$ represents the amplitude of the frequency $(0,0)$, hence we do not initialize it in $\omega$.

\begin{figure}[!h]
\centering
    \includegraphics[width=0.65\textwidth]{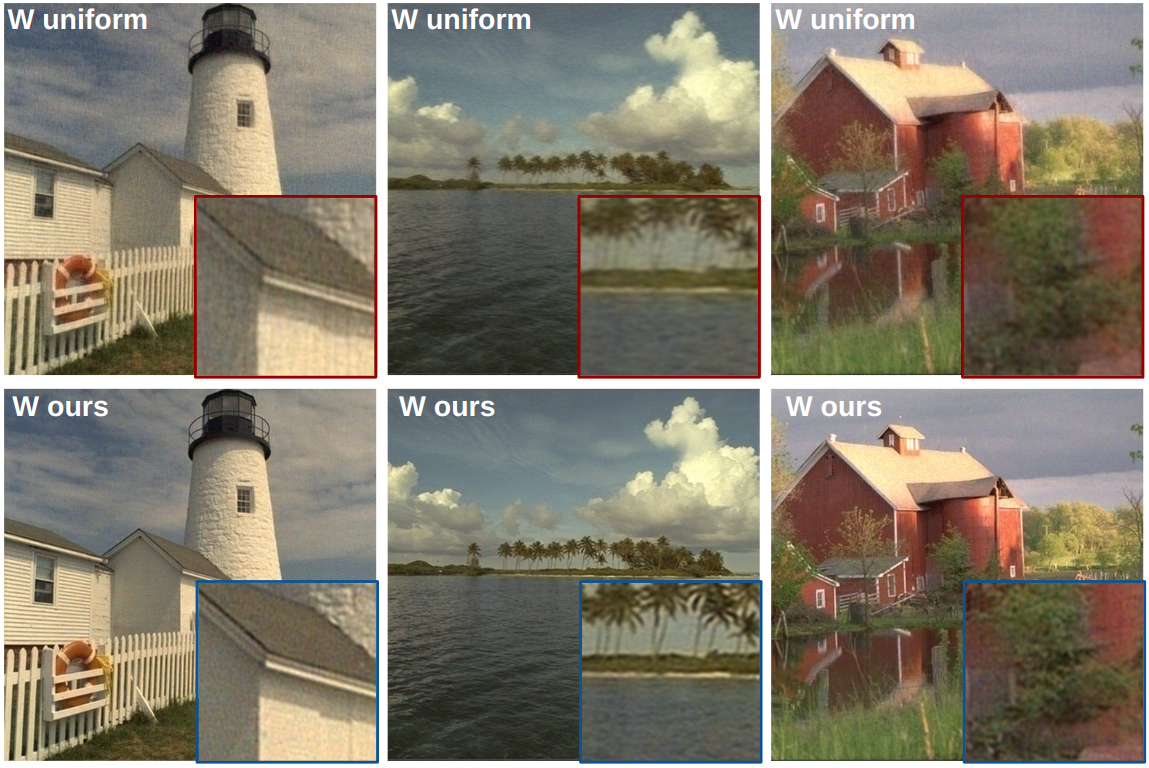}
    \vspace{-0.3cm}
    \caption{Comparison between INRs with different initializations for $\W$.
    The row above/below shows the reconstruction of a network with $\W$ initialized as in \cite{sitzmann2020implicit}/Sec~4.2.
    We trained during $10$ epochs.
    The blue and red squares present a zoom-in of the image center.}
    \label{fig: W_init}
    % \vspace{-0.4cm}
\end{figure}

In Sec~4.2 of the main paper we proposed an initialization for the hidden matrix $\W$ based on the bound values. The following experiment shows that such an initialization may grant faster convergence.
In Fig~\ref{fig: W_init}, we initialize the INRs with size $m=416$ and $n=1024$, bandlimit $\gt{b}=82$, and bounds $\gt{c}_\textbf{L}=1.0$, $\gt{c}_\textbf{H}=0.2$. We train the networks during $10$ epochs and fit an image with resolution $1024^2$.
Observe that our initialization for $\W$ (below) offers a better reconstruction than the uniform initialization (above). This can also be observed in the zoom-ins of images, where our method presents more details.

\subsection{Learned bounds}
\label{a-initialization-hidden-layer}

\begin{figure}[!h]
    \centering
    \includegraphics[width=0.73\textwidth]{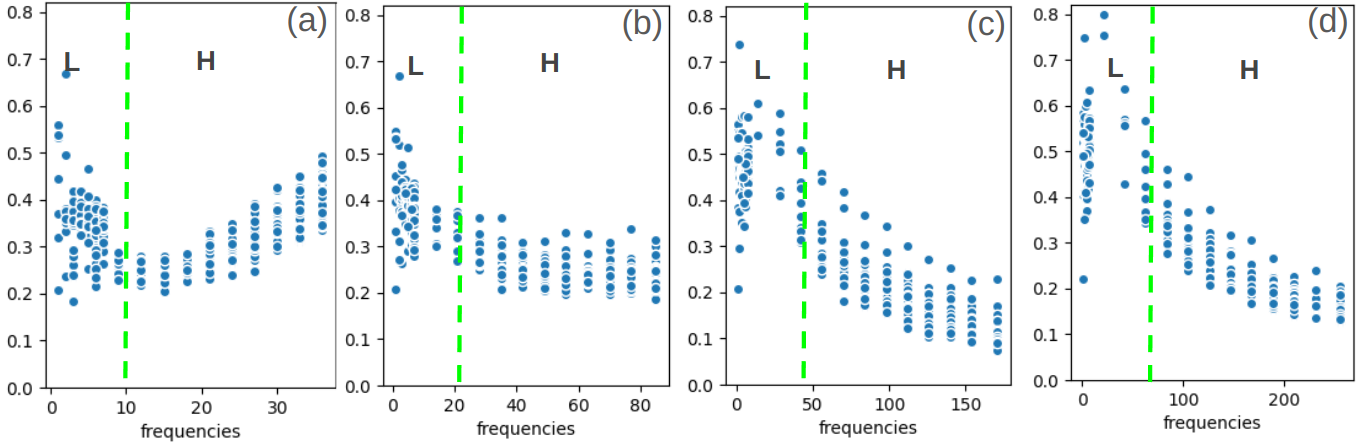}
    \vspace{-0.3cm}
    \caption{Ablation of the learned bounds of each column. The $x$-axis corresponds to the maximum coordinate (in absolute) of the frequency $\omega_j$, while the $y$-axis shows the trained bound $\gt{c}_j$ of $\W$'s $j$-th column. Thus, the blue points in the diagram represent $(\max(|\omega_j|), \gt{c}_j)$.}
    \vspace{-0.2cm}
    \label{fig: learn_bounds_ablation}
\end{figure}

Sec~4.2 of the main paper introduced an architecture that enables learning the bounds during training.
Here, we study its behavior as the sampling of input frequencies vary.
We compare sinusoidal INRs of size $m=n=416$, initializing the learned bounds as $0.5$ and varying the bandlimit between $\gt{b} = 41, 85, 171, 256$. The lower frequency limit is set to $\gt{l}=\nicefrac{\gt{b}}{4}$.
Fig~\ref{fig: learn_bounds_ablation} shows the bounds trained over $400$ epochs on a $512^2$ resolution image. Each plot includes a green dashed vertical line that splits the low (\textbf{L}) and high (\textbf{H}) frequencies.
Figs~\ref{fig: learn_bounds_ablation}(a)-(b) show the learned bounds when initializing using small $\gt{b}$, leading to similar bounds across all frequencies.
On the other hand, Figs~\ref{fig: learn_bounds_ablation}(c)-(d) consider higher values of $\gt{b}$. Here, low/high frequency bounds increase/decrease, reducing noise generation as bigger bounds may imply in higher multiples of the input frequencies.
This behavior aligns with our claims that such bounds serve as a mechanism of spectral control.

\subsection{Representational capacity of layer composition}
\begin{figure}[!h]
     \centering
    \includegraphics[width=0.45\columnwidth]{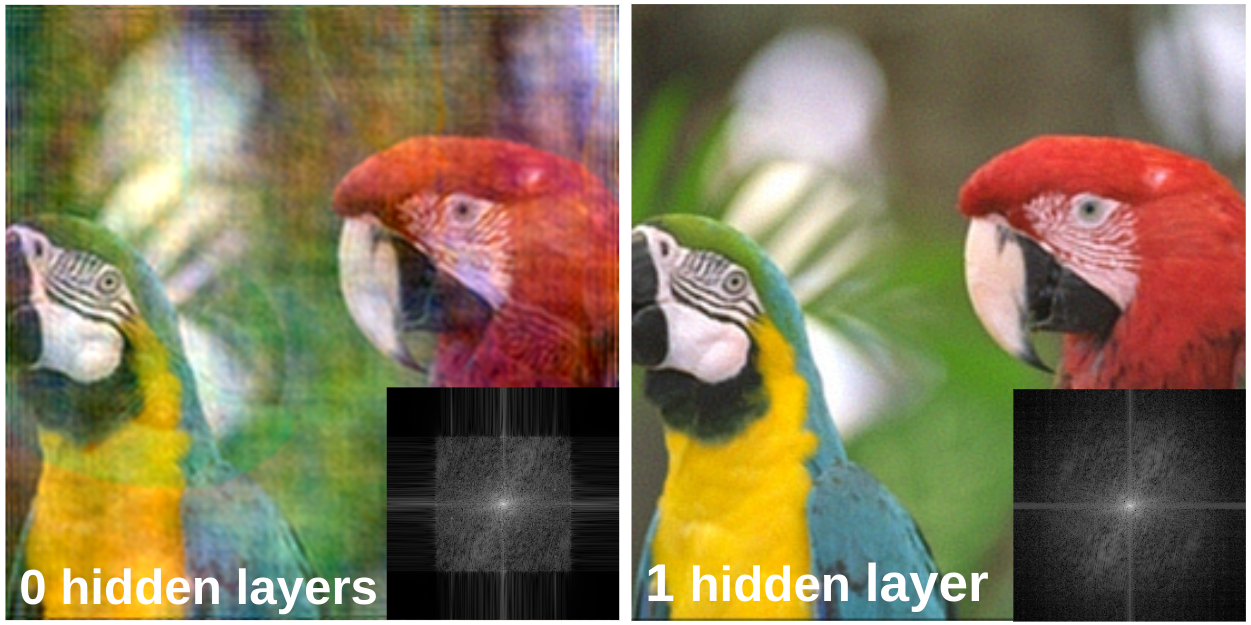}
    \vspace{-0.2cm}
    \caption{Comparison between INRs for fitting images, with zero and one hidden layers. Training with one hidden layer is significantly faster ($50$s vs $13$m), uses half the parameters ($33842$ vs $67803$), and produces higher quality results ($34.4$dB vs $21.1$dB).
    }
    \label{fig: freq_fabric}
\end{figure}

Theorem~\ref{thm: thm-1} states that a sinusoidal INR with a single hidden layer and input frequencies $\omega$ can represent a signal using an infinite number of frequencies $\beta_\textbf{k}=\dotprod{\textbf{k}, \omega}$. 
Here, we illustrate how the composition of layers enables a more compact and expressive representation by generating additional frequencies.
Specifically, Fig~\ref{fig: freq_fabric} compares the representation capacity of a wide but shallow network and a narrower, deeper network. 
For the shallow case, we use a INR with no hidden layers $f(\textbf{x})=\textbf{C}\sin(\omega\textbf{x}+\varphi)+e$ where $\omega\in\frac{2\pi}{p}\Z^{11300\times2}$. 
Training in this case optimizes the amplitudes $\textbf{C}$ of the frequencies $\omega$.
For the deeper case, we train a network with a hidden layer composition, using a configuration of $m=120$ and $n=239$.

Theorem~\ref{thm: thm-1} then implies that $f$ has as an expansion with amplitudes determined by the hidden weights. Thus, training a deeper sinusoidal INR also fits the amplitudes of a sum of sines, but with a much wider number of frequencies.
Indeed, Fig~\ref{fig: freq_fabric} shows that this network not only converges faster but also achieved better quality with half the number of parameters.

Next, we extend our ablations to include deeper sinusoidal MLPs. Specifically, we consider MLPs of two hidden layers with $256$ neurons each and train them on Kodak dataset images with a resolution of $512^2$. We set the bounds as $\gt{c}_\textbf{L}=1.0$, $\gt{c}_\textbf{H}=0.3$, and $\gt{c}_\textbf{2}=0.05$, where the last one corresponds to the bound of the hidden matrix $\W^2$. 
Table~\ref{tab: tuner} compares our initialization (2rd column) of the input frequencies with uniform initialization (1rd column), using $\gt{b}\!\!=\!\!128$, $\gt{l}\!\!=\!\!10$, $70\%$ of $\omega$ set as lower frequencies, and $200$ training steps.  
Additionally, we include a comparison with our full method (3rd column). 
This shows that TUNER is also effective for deeper MLPs.
\begin{table}
\centering
\begin{tabular}{l|ccc}  
\hline  
          & \color{red!60}{rand init}       & \color{teal!60}{our init}  & \color{blue!60}{TUNER}        \\ \hline  
% \texttt{RGB} & 23.54 & 32.43 & 30.81 \\  
\texttt{RGB}  & 16.86 & 28.47 & \textbf{28.58}  \\  
grad          & 17.67 & 24.95 & \textbf{25.42}  \\ \hline  
\end{tabular}  
\vspace{-0.2cm}
\caption{Experiments with deeper MLPs. We compare {\color{teal!60}{our initialization}} of the input freq. with {\color{red!60}{uniform}} and our full method ({\color{blue!60}{\small 3rd column}}). }
    \label{tab: tuner}
\end{table}

\subsection{Additional comparisons}
\begin{figure}[h!]
% \begin{wrapfigure}[16]{r}{0.7\columnwidth}
\centering
    \vspace{-0.3cm}
    \includegraphics[width=0.7\columnwidth]{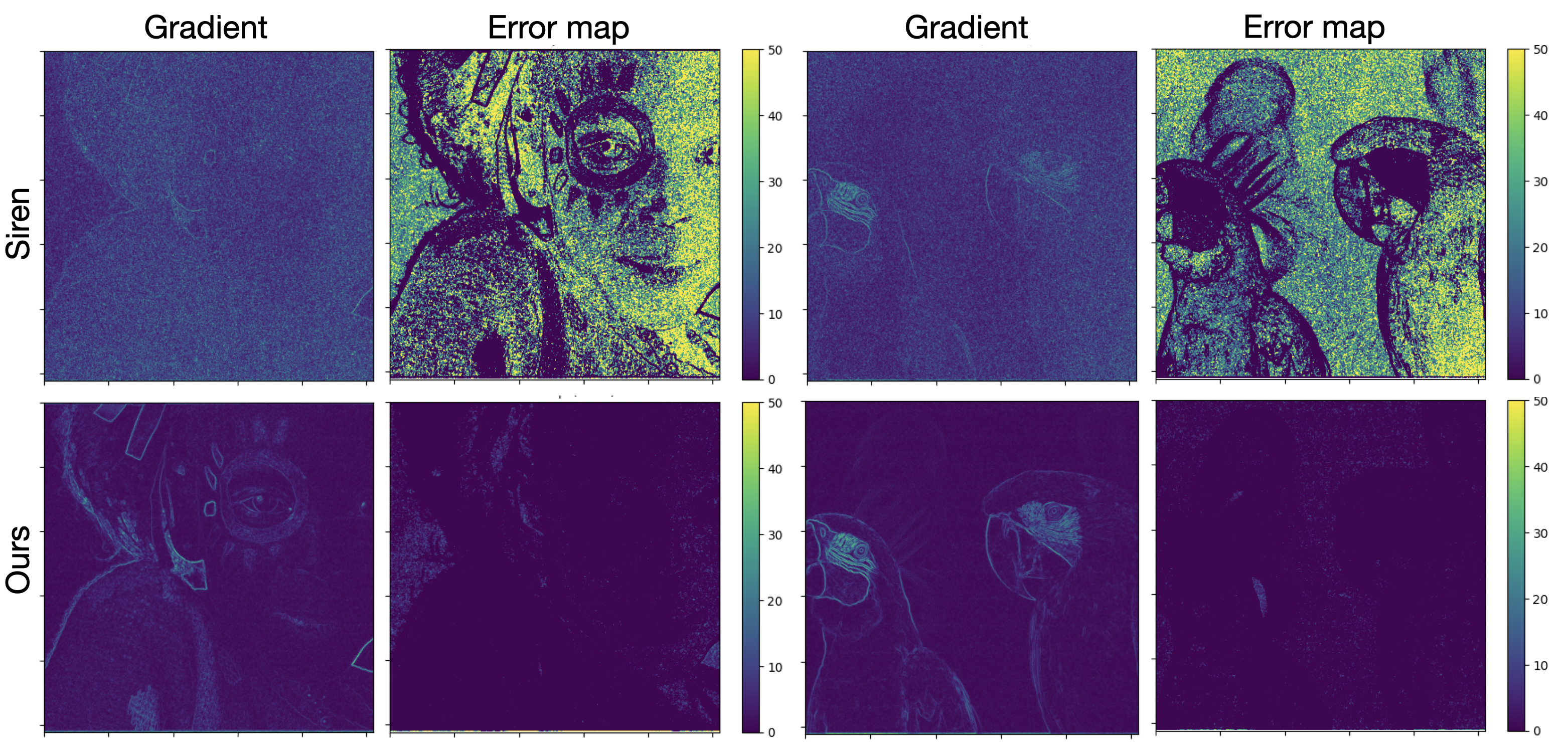}
    \vspace{-0.2cm}
    \caption{Gradient visualization of experiments in Table 3. The error map shows that while SIREN fits well only the higher frequencies, ours fit both.}
    \label{f-pipeline}
\end{figure}

% Table~\ref{tab:all-methods} offers an overview of their characteristics.
%
% \begin{table}[h!]
% % \addtolength{\tabcolsep}{-4pt}
% \centering
% % \footnotesize
% % \vspace{-0.3cm}
%  {\begin{tabular}{l|ccc}
% \cline{2-4}
%  & \textbf{Initialize} & \textbf{Activation} & \textbf{Filter} \\ \hline
% SIREN & $\omega, \W$ & $\sin$ & - \\
% FFM & $\omega$ & $\sin / \cos / \texttt{ReLU}$ & - \\
% BACON & Base & $\sin$ & Box filter \\
% BANF & {standard \texttt{ReLU}} & $\texttt{ReLU}$ & Linear/sinc \\
% FINER & $\textbf{b}$ & $\sin((1+|\cdot|)\cdot)$ & - \\
% TUNER & $\omega, \W$ & $\sin$ & Bounding \\ \hline
% \end{tabular}
% \caption{ {INR methods related to TUNER. Here, ``Standard" initialization refers to \citet{sitzmann2020implicit} approach to train sinusoidal INRs, ``Default" to Pytorch~\cite{paszke2017automatic} initialization while ``N/A" in the 4th column indicates the method doesn't use any mechanism to bandlimit the INR.}}
% \label{tab:all-methods}}
% \end{table}
This section provides additional comparisons between different approaches to signal fitting using sinusoidal INRs and their variations.  
%
% First, we compare SIREN, BACON, and TUNER in terms of their bandlimiting capacity, using SIREN as the baseline architecture. 
 Figure~\ref{f-pipeline} presents the gradient and error maps of SIREN and TUNER reconstructions after 3000 training epochs. We use networks of size \( m = n = 416 \) with \( \gt{b} = 171 \), training them on images of resolution \( 512^2 \) while supervising with 90\% of the available samples.

Note that compared to SIREN, our gradient has no notable noise in both images.
In particular, note that the error maps of SIREN have low error around regions of high detail (such as the silhouette of the girl, or near the eyes of the macaws) while regions of low detail have high error (like the background of the macaws or the face of the girl).
This indicates that SIREN represented well the areas with high frequency content but those as propagated as noise in regions of low frequency content.
Conversely, our error maps show that TUNER was able to represent well most of the image, tuning the frequencies to the spectral content of each region.

% \pagebreak

\begin{figure}[h!]
    \centering
    \includegraphics[width=0.85\textwidth]{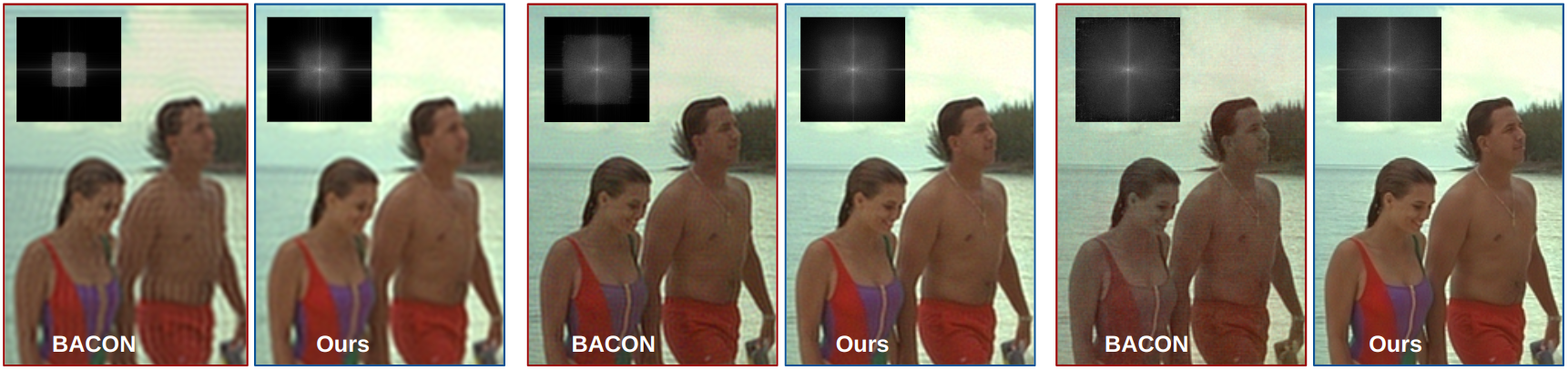}
    \vspace{-0.3cm}
    \caption{Comparison between BACON  (red) and our method (blue), trained with different bandlimits $\gt{b}=85, 171, 256$. We observe that BACON uses a box filter, generating ringing artifacts (first image). In contrast, our method resembles a soft filter, improving quality.}
    \label{fig: bacon_ours}
    \vspace{-0.3cm}
\end{figure}

\paragraph*{} Now, we show an extended version of the bandlimit control experiment from Fig~1 (main paper).
Specifically, we compare BACON (red) and TUNER (Ours, in blue), trained using different bandlimits for the network ($85, 171, 256$). 
In Fig~\ref{fig: bacon_ours}, we observe that for the low bandlimit (leftmost two), BACON generates visible ringing artifacts. In contrast, our method resembles a soft filter preserving the smoothness of the reconstruction.
Conversely, when increasing the bandlimit of the spectrum, the reconstruction of BACON distorts color, while ours improves reconstruction.

\begin{figure}[h!]
\centering
    \includegraphics[width=0.75\columnwidth]{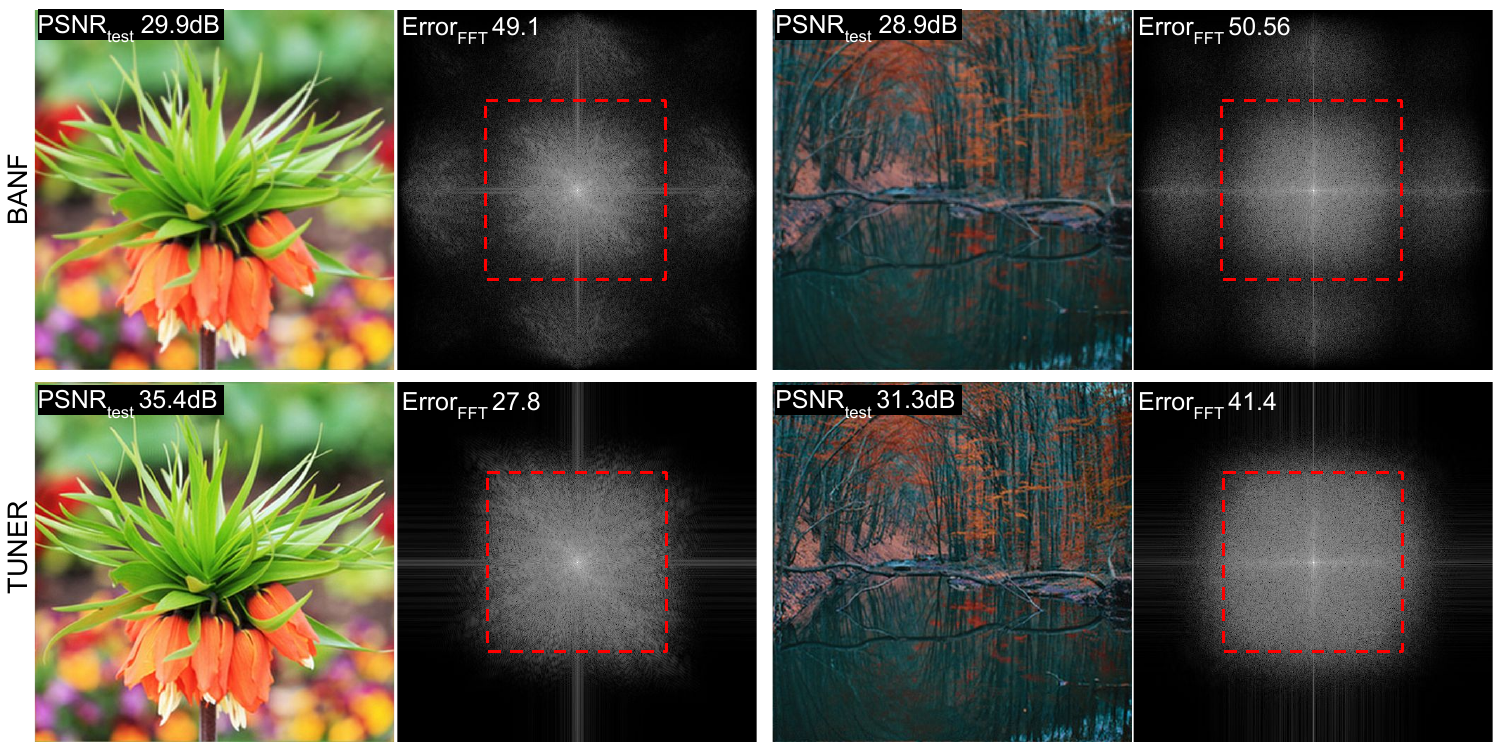}
    \caption{ {Comparison between TUNER and BANF in the signal (1st and 3rd column) and spectral (2nd and 4th columns) domains. Note that our reconstruction offers more details while restricting the appearance of frequencies to the band (red square) akin to a soft filter. Conversely, BANF exhibits artifacts in the spectrum, even though its representation appears blurrier.}}
    \label{f-banf}
    \vspace{-0.2cm}
\end{figure}

\paragraph*{TUNER and BANF.}
{We present a numerical comparison with BANF on the DIV2K dataset \cite{Agustsson_2017_CVPR_Workshops}, rescaling the images to a resolution of $256^2$. We compare different bands ($\gt{b}=64$ and $\gt{b}=256$) for the spectrum and measure the PSNR in a resolution of $512^2$.
To assess the bandlimiting capacity of each network, we consider the amplitudes $\alpha_\textbf{k}$ of frequencies $\dotprod{\textbf{k}, \omega}$ outside of the band $\mathcal{B}$ to be noise. Then, we define the bandlimit error as:}
$$
 {\text{Error}_{\text{FFT}} = \sum_{\dotprod{\textbf{k}, \omega}\notin \mathcal{B}}|\alpha_\textbf{k}|}.
$$
{For each level of detail in BANF, we initialized and trained a new TUNER INR from scratch. The epochs used were adjusted so that our training time was equivalent to BANF's, since their epochs took longer to train. The quantitative results are summarized in  Table~\ref{tab:banf}.
Observe that TUNER outperforms BANF on both metrics for all bandlimits, even when it is not trained over the residuals of previous resolutions.
\begin{table}[h!]
\centering
% \footnotesize
\begin{tabular}{l|cc|cc}
\hline
      & \multicolumn{2}{c|}{PSNR$\uparrow$} & \multicolumn{2}{c}{$\text{Error}_\text{FFT}$ $\downarrow$} \\ \hline
Band &  64     &  256    &  64       &  256      \\ \hline
BANF  & 22.96       & 29.59       & 30.71         & 50.17         \\
TUNER & \textbf{26.31}       & \textbf{32.65}       & \textbf{18.97}         & \textbf{36.16} \\
\hline
\end{tabular}
\caption{ {Comparison of TUNER and BANF in signal quality and bandlimiting error $\text{Error}_\text{FFT}$ with bands $64$ and $256$.
% We define the bandlimit error  as the sum of amplitudes (in absolute value) of frequencies outside the designated band. 
Our method has better quality ($\approx 3$dB) and reduces the appearance of frequencies outside the band.}}
\label{tab:banf}
% \end{table}
\end{table} 

We also present a qualitative comparison with BANF, showing two reconstructions with a resolution of $512^2$ trained with a bandlimit of $128$, under the same conditions as in  Table~\ref{tab:banf}. 
Figure~\ref{f-banf} shows that our method preserves more details (with at least a $3$dB of improvement) while enforcing a smooth filtering outside the band (red square).
In contrast, BANF exhibits many frequencies outside the specified bandlimit, which is reflected in the higher $\text{Error}_\text{FFT}$ values.
}

\begin{figure}[h!]
    \centering
    % \vspace{-0.4cm}
    \includegraphics[width=0.75\columnwidth]{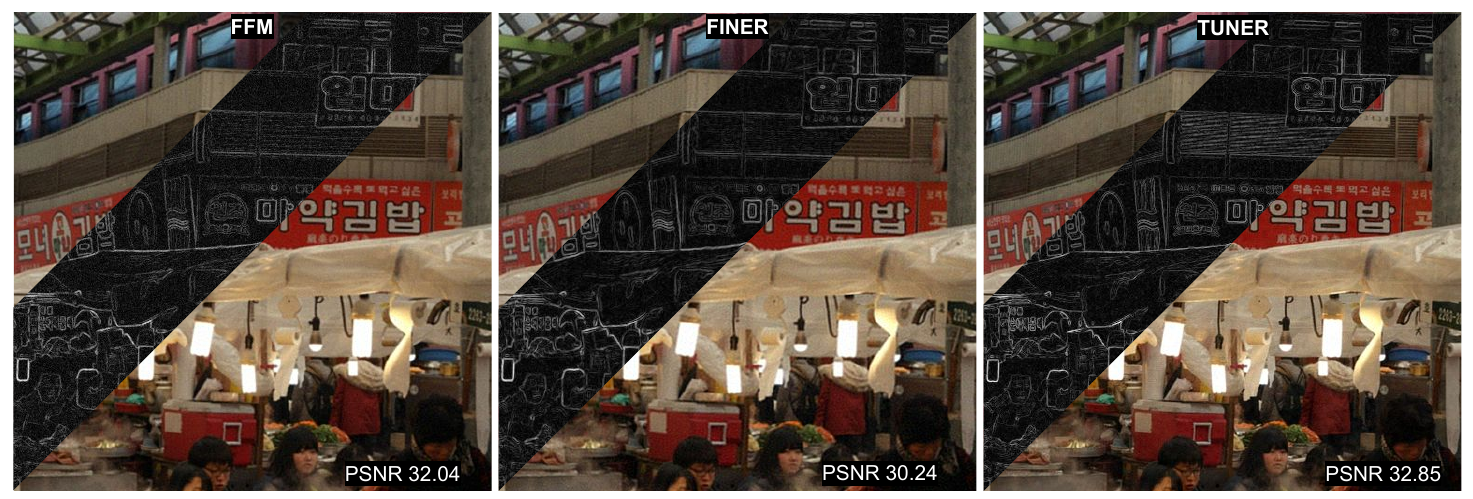}
    \caption{ {Comparison between Fourier Feature Mapping (FFM), FINER and TUNER in the \texttt{RGB} and gradient (grayscale band) domains after training for 1000 epochs. Observe that FFM reconstruction has a signal quality comparable to ours, but their gradients are noisier, while FINER presents a smoother gradient but fails to reconstruct more detailed regions.}}
    \label{fig:finer_vs_tuner}
\end{figure}

\paragraph*{Compare with FINER \& Fourier Feature Mapping (FFM).}
We present a comparison between Fourier Feature Mapping (FFM), FINER, and TUNER on the DIV2K dataset with images of resolution \( 512^2 \).
All networks are configured with a single hidden layer of size \( m = n = 256 \) and their corresponding initializations. For FINER, we use \( \gt{b} = 85 \) and \( \textbf{b} \sim \mathcal{U}(-1, 1) \). The TUNER INR has a period of 3 and bounds \( \gt{c}_\textbf{L} = 1.0 \), \( \gt{c}_\textbf{H} = 0.6 \).
All networks are trained for 5000 epochs using the Adam optimizer with a learning rate of \( 5 \times 10^{-4} \).
As shown in Table~\ref{tab:tunerVSfiner}, TUNER outperforms both methods with at least a 2~dB improvement. We also observe a qualitative improvement in Figure~\ref{fig:finer_vs_tuner}, where our method achieves comparable quality to previous works while enhancing the reconstruction of higher-order information.
\begin{table}[!h]
    \centering
    \vspace{-0.2cm}
    \begin{tabular}{l|ccc}
        \hline
        epochs & FFM          & FINER        & TUNER        \\ \hline
        1000     & 29.42  & 30.23 & \textbf{32.14} \\
        5000    & 31.19  & 31.00 & \textbf{33.16} \\ \hline
    \end{tabular}
    \caption{ {Comparison between Fourier Feature Mapping (FFM), FINER and TUNER when training during 1000 and 5000 epochs.}}
    \label{tab:tunerVSfiner}
\end{table}

% \pagebreak
\bibliographystyle{abbrvnat}
\bibliography{refs}